\documentclass[10pt,journal,compsoc]{IEEEtran}
\ifCLASSOPTIONcompsoc
  \usepackage[nocompress]{cite}
\else
  \usepackage{cite}
\fi
\ifCLASSINFOpdf
\else
\fi
\usepackage{amsmath,amsfonts}
\usepackage{algorithmic}
\usepackage{array}
\usepackage[caption=false,font=normalsize,labelfont=sf,textfont=sf]{subfig}
\usepackage{textcomp}
\usepackage{stfloats}
\usepackage{url}
\usepackage{verbatim}
\usepackage{graphicx}

\usepackage{algorithm}
\usepackage{algorithmic}

\usepackage{newfloat}
\usepackage{listings}

\usepackage{multirow}
\usepackage{diagbox}
\usepackage{booktabs}
\usepackage{subfig}
\usepackage{bbding}
\usepackage{float}
\usepackage[inline]{enumitem}
\usepackage{amsmath}
\usepackage{makecell}

\usepackage{xcolor}

\begin{document}
%
\title{Label-Free Multivariate Time Series Anomaly Detection}
%
%
%
%

\author{Qihang Zhou,~\IEEEmembership{Student Member,~IEEE,}
        Shibo	He,~\IEEEmembership{Senior Member,~IEEE,}
        Haoyu Liu,\\
        Jiming	Chen,~\IEEEmembership{Fellow,~IEEE,}
        and Wenchao	Meng,~\IEEEmembership{Senior Member,~IEEE}
   
\IEEEcompsocitemizethanks{
\IEEEcompsocthanksitem Part of this work was published in Proceedings of the AAAI'23~\cite{zhou2023detecting}.

\IEEEcompsocthanksitem Q. Zhou, J. Chen, S. He, and W, Meng are with the State Key Laboratory of Industrial Control Technology, Zhejiang University, Hangzhou, Zhejiang, 310027, China. E-mail: \{zqhang, cjm, s18he, wmengzju\}@zju.edu.cn. \textit{Corresponding author: Shibo He}

\IEEEcompsocthanksitem H. Liu is with the State Key Laboratory of Industrial Control Technology, Zhejiang University, Hangzhou 310027, China, and also with Fuxi AI Lab, NetEase Games, Hangzhou 310052, China (e-mail: liuhaoyu03@corp.netease.com).

Copyright © 20xx IEEE. Personal use of this material is permitted. However, permission to use this material for any other purposes must be obtained from the IEEE by sending an email to pubs-permissions@ieee.org.
}}

\markboth{Journal of \LaTeX\ Class Files,~Vol.~14, No.~8, August~2015}%
{Shell \MakeLowercase{\textit{et al.}}: Bare Demo of IEEEtran.cls for Computer Society Journals}
%



\IEEEtitleabstractindextext{%
\begin{abstract}
Anomaly detection in multivariate time series has
been widely studied in one-class classification (OCC) setting. The training samples in this setting are assumed to be normal. In more practical situations, it is difficult to guarantee that all samples are normal. Meanwhile, preparing a completely clean training dataset is costly and laborious. Such a case may degrade the performance of OCC-based anomaly detection methods which fit the training distribution as the normal
distribution. To overcome this limitation, in this paper, we propose MTGFlow, an unsupervised anomaly detection approach for \underline{M}ultivariate \underline{T}ime series anomaly detection via dynamic \underline{G}raph and entity-aware normalizing \underline{Flow}. MTGFlow first estimates the density of the entire training samples and then identifies anomalous instances based on the density of the test samples within the fitted distribution. This relies on a widely accepted assumption that anomalous instances exhibit more sparse densities than normal ones, with no reliance on the clean training dataset. However, it is intractable to directly estimate the density due to the complex dependencies among entities and their diverse inherent characteristics, not to mention detecting anomalies based on the estimated distribution. 
In order to address these problems, we utilize the graph structure learning model to learn interdependent and evolving relations among entities, which effectively captures the complex and accurate distribution patterns of multivariate time series. In addition, our approach incorporates the unique characteristics of individual entities by employing an entity-aware normalizing flow. This enables us to represent each entity as a parameterized normal distribution. Furthermore, considering that some entities present similar characteristics, we propose a cluster strategy that capitalizes on the commonalities of entities with similar characteristics, resulting in more precise and detailed density estimation. We refer to this cluster-aware extension as MTGFlow\_cluster. Extensive experiments are conducted on six widely used benchmark datasets, in which MTGFlow and MTGFlow\_cluster demonstrate their superior detection performance.

\end{abstract}

}

\maketitle

\IEEEdisplaynontitleabstractindextext

%
\IEEEpeerreviewmaketitle

\IEEEraisesectionheading{\section{Introduction}\label{sec:introduction}}

%
%
%
%

\begin{figure}
  \centering
    \subfloat[OCC-based methods]{\includegraphics[width=0.47\columnwidth]{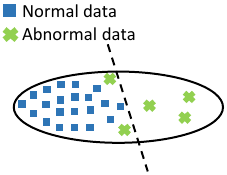}%
    \label{fig: OCC-based methods}}
    \hfil
    \subfloat[Density-based methods]{\includegraphics[width=0.5\columnwidth]{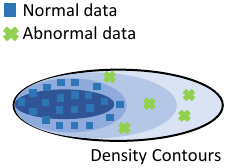}%
    \label{fig: Density-based methods}}
 \caption{Schematic diagram of OCC-based and density-based methods on a dataset with mixed normal and abnormal data. \textbf{Left}: OCC-based methods are vulnerable to noisy labels, leading to inaccurate decision boundaries. \textbf{Right}: Unlike OCC-based methods, density-based methods estimate the density of the test samples, where anomalies are located in low-density regions. Note that the deeper the color, the higher the density.}
 \label{fig:auroc_curves}
\end{figure}

\IEEEPARstart{M}{ultivariate} time series (MTS) are commonly found in various contexts, such as smart factories where data is produced by multiple devices, and smart grids where monitoring data is generated by various sensors~\cite{he2022collaborative}. Anomalies in MTS refer to unusual patterns or behaviors that occur at a particular time or over a specific period of time~\cite{chandola2010anomaly, gaddam2007k, yang2022detecting, pu2022security}. Previous methods mainly center on developing one-class classification (OCC) models, which rely solely on normal data to detect anomalies~\cite{OC-SVM,su2019robust,chen2021daemon, deng2021graph,xu2021anomaly, zhang2019deep}. The fundamental assumption of them is that the training dataset with all normal samples can be easily obtained~\cite{ruff2021unifying, Zhou9940966}. However, the premise that the training dataset contains all normal samples may not always hold in real-world applications~\cite{zhou2023anomalyclip, goodge2021lunar, zhang2019online, zong2018deep, qiu2022latent, liu2022time, zhang2023trid}. This leads to noisy training datasets with a mixture of normal and abnormal data instances. Using these training datasets will result in overfitting of the model to noisy labels~\cite{zhang2021understanding}. As shown in Fig.~\ref{fig: OCC-based methods}, this degrades the performance of those OCC-based methods~\cite{wang2019effective, huyan2021unsupervised}. Therefore, it is rewarding to develop unsupervised MTS anomaly detection methods based on the dataset with absolute zero known labels.

Density estimation is a promising approach for unsupervised anomaly detection because they do not depend on the assumption that training datasets are all normal. As presented in Fig.~\ref{fig: Density-based methods}, it detects an abnormal sample based on the widely accepted hypothesis that abnormal instances exhibit sparse densities than the normal, i.e., abnormal samples are typically in the low-density regions while normal samples are in the high-density regions~\cite{gupta2013outlier, Guansong, Ruoying}. Previous works have attempted this strategy for time series prediction, and the key challenge lies in the accurate density estimation of the distribution. 
In~\cite{salinas2020deepar, rasul2021autoregressive, feng2022multi}, time series density is modeled as the parameterized probability distribution,
while it is still challenging to model a more complex data distribution~\cite{rasul2020multivariate}.

More recently, GANF~\cite{dai2021graph} explores the density estimation to tackle MTS anomaly detection task. In their design, the static directed acyclic graph (DAG) is leveraged to model intractable dependence among multiple entities, and normalizing flow is employed to estimate an overall distribution for all entities together~\cite{dinh2016density, papamakarios2017masked}. Although GANF has achieved state-of-the-art (SOTA) results, it may suffer from two drawbacks. First, rather than a static inter-relationship, in real-world applications, the mutual dependencies among entities could be evolving. For instance, in a water treatment plant, when the motorized valve is turned on, it causes the water level to rise. And when the water level exceeds a certain threshold, the motorized valve is closed. These interactions between the actuator of the motorized valve and the water level sensor are strongly coupled. Meanwhile, due to different work conditions, the water level needs to rise for storing water or decrease for draining water. Such relationships are not set in stone, and thus the static graph structure fails to capture this evolving feature. 
\textbf{Question 1: Since dependencies among multi-entities are mutual, is there a better choice to model complicated dependencies than the probabilistic model DAG?} Second, entities usually have diverse working mechanisms, leading to diverse sparse characteristics when anomalies occur. And, GANF projects all entities into the same distribution. \textbf{Question 2: Individual anomalous time series from different entities present diverse sparse characteristics. Is it suitable to map all these individual time series to the same latent space (i.e., normal Gaussian distribution)?}

In this paper, we propose MTGFlow and  MTGFlow\_cluster, unsupervised
anomaly detection methods for MTS anomaly detection, to answer the above two questions. First, considering the evolving relations among entities, we introduce graph structure learning to model these changeable interdependencies. To learn the dynamic structure, a self-attention module~\cite{vaswani2017attention} is plugged into our model for its superior performance on quantifying pairwise interaction. Second, we introduce an entity-aware normalizing flow to handle the diverse inherent characteristics among entities. This approach enables us to model entity-specific density estimation, allowing each entity to be associated with a unique target distribution. Consequently, we can estimate the densities of various entities independently. Moreover, the characteristics of entities are strongly connected to their positions in common scenarios. In industrial IoT, sensors within the same production line or workshop have common characteristics due to their exposure to the same industrial processes and operating conditions~\cite{rodrigues2008hierarchical}. For example, temperature and humidity sensors monitor the same area~\cite{gaddam2007k, Li2018robust}. Similarly, the common wind speed sensors in wind turbines exhibit similar wind speeds and variations in the same regions~\cite{liu2022wind}. In the domain of transportation, traffic flow sensors along the same stretch of highway or road will exhibit similar patterns because they are influenced by similar traffic conditions and weather~\cite{aboah2021vision}. Identifying the common information makes for fine-grained representation learning for downstream tasks~\cite{paparrizos2015k, NEURIPS2019_1359aa93}. Motivated by this, we propose to cluster these entities into different groups. Each group is assigned to the same target distribution while also preserving the distinctiveness among different groups. This results in the creation of MTGFlow\_cluster. Such a clustering strategy enables us to efficiently handle entities with similar properties and better model their respective density distributions. Incorporating these proposed technologies provides a powerful framework for tackling the complicated characteristics among diverse entities.
However, this fine-grained density estimation of MTGFlow and MTGFlow\_cluster multiplies memory overhead as the number of entities/clusters increases. We share entity/cluster-specific model parameters to reduce model size. As a result, MTGFlow and MTGFlow\_{cluster} achieve more fine-grained density estimation without extra memory consumption. 

Finally, all modules of MTGFlow and MTGFlow\_{cluster} are jointly optimized for overall performance, and maximum log likelihood estimation (MLE) is used to train them. We explore MTGFlow and its extension in various settings including unsupervised and OCC settings. By comparing empirical results of these two settings, MTGFlow and MTGFlow\_{cluster} are found to be robust to anomaly contamination and adapted to more general scenarios. Experiments are conducted on six public datasets to demonstrate the effectiveness of MTGFlow and MTGFlow\_{cluster}, which make progress over the state-of-the-art (SOTA) methods~\cite{ruff2019deep, ruff2018deep, sabokrou2020deep, goyal2020drocc, audibert2020usad, zong2018deep, dai2021graph}, particularly outperforming 5\% on SWaT. Besides, we attempt to apply MTGFlow to univariate series (i.e., UCR dataset~\cite{wu2021current}), and it still attains superior results. 

Our contributions are summarized as follows:

\begin{itemize}

\item We propose MTGFlow and MTGFlow\_cluster for unsupervised MTS anomaly detection. It essentially enables anomaly localization and interpretation without any labels.

\item We model the complicated dependencies among entities into the dynamic graph, capturing the complex and evolving mutual dependencies among entities.

\item Aiming at different sparse characteristics existing in individual entities, entity-aware normalizing flow is introduced to produce entity-specific density estimation. 
\item 
Considering the characteristics of entities are strongly connected to their positions, we propose MTGFlow\_{cluster}, which exploits cluster-aware normalizing flow to capitalize on the commonality of these entities and keep the uniqueness among entities with different characteristics.

\item Experiments on six datasets with seven baseline methods demonstrate the superiority of MTGFlow and MTGFlow\_cluster, which outperform the SOTA methods. 

\end{itemize}

It is worth mentioning that a conference version of this research was presented in~\cite{zhou2023detecting}. The conference paper does not involve any priors. To consider the diverse characteristics of entities, MTGFlow maps all entities to unique distributions for anomaly detection. In reality, many entities in industrial or smart grids usually present similar characteristics~\cite{moghaddass2017hierarchical, ghafouri2020detection}. For example, water level sensors are used to detect the water level in the same area but in different locations. The cluster information is readily available, and mapping similar characteristics into the different distributions is not an optimal choice. Therefore, we propose MTGFlow\_cluster to perform more fine-grained density estimation, which not only considers the diverse characteristics among entities but also introduces a clustering strategy that capitalizes on the commonalities of entities with similar characteristics. Given the cluster prior, this approach has the potential to further improve anomaly detection performance.
Moreover, we supplement the experiments about the OCC setting and univariate time series for the overall evaluation. In addition, more illustrations and visualizations are provided to make a comprehensive analysis of our proposed methods.

\section{Related Work}
In this section, we introduce recent work in the field of time series anomaly detection, with a particular focus on graph structure learning and normalizing flow techniques.
\subsection{Time Series Anomaly Detection}

Anomaly detection for time series is a classical research topic, which has been extensively investigated under OCC setting~\cite{chalapathy2019deep}. Temporal correlation is one of the most important features of time series~\cite{hundman2018detecting}. Compared with the normal, anomalous time points and sequences often present unusual temporal correlations. To model the distribution of normal time series, DeepSVDD~\cite{ruff2018deep} maps training data into preset hypersphere, assuming that anomalous data lie outside this space during the test. EncDecAD~\cite{malhotra2016lstm} leverages LSTM~\cite{hochreiter1997long} to extract sequence features and designs the reconstruction task to detect anomalies. USAD~\cite{audibert2020usad} and DAEMON~\cite{chen2021daemon} use adversarial learning to promote reconstruction quality. 
Considering the much more complex temporal dependence of MTS, OmniAnomaly~\cite{su2019robust} constructs informative stochastic representations for more robust performance. More recent works~\cite{xu2021anomaly} and~\cite{tuli2022tranad} utilize Transformer~\cite{vaswani2017attention} for anomaly detection, leaning on the superiority modeling capacity of the self-attention mechanism for long-range relations. 

However, all these works are based on the assumption that a sufficient training dataset with all normal instances can be acquired, which is very hard for real-world applications since each instance should be manually checked carefully. Additionally, once there exist abnormal instances in the training data, the performance of these OCC-based detection methods could be severely degraded~\cite{wang2019effective, huyan2021unsupervised}. Therefore, instead of fitting the distribution of normal training datasets, Dai and Chen propose GANF~\cite{dai2021graph} to detect MTS anomalies in an unsupervised manner. Inspired by them, we propose MTGFlow to facilitate the learning capacity and improve detection performance.

\subsection{Graph Structure Learning for Anomaly Detection}
Given a graph structure, graph neural networks ~\cite{kipf2016semigcn, xu2018powerful} have achieved great success in modeling intrinsic structure patterns.  Wang~\emph{et al.}~\cite{pmlr-v162-tang22b} extend the OCSVM to graph anomaly detection, combining the powerful representation ability of GNN. Tang~\emph{et al.}~\cite{wang2021one} analyze the anomaly detection from the graph spectral domain and find graph anomalies lead to a shift of spectral energy distributions. They generate band-pass filters to capture this energy shift and further perform anomaly detection. However, In real
scenarios, the prior graph structure is hard to provide. Therefore, it is important to learn the underlying graph structure via the graph embedding itself~\cite{velivckovic2017graphgat}. GDN~\cite{deng2021graph} learns a directed graph via node embedding vectors. According to the cosine similarity of embedding vectors, top-K candidates of each node are considered to have dependencies. They rely on either the label supervision or normal training dataset. In the field of unsupervised anomaly detection of MTS, some recent work attempts to explore this area. GANF~\cite{dai2021graph}  uses a DAG to model the relationships between multiple sensors and then learns the structure of the DAG through continuous optimization while applying a simplified constraint to facilitate backward propagation. Our work, namely MTGFlow and MTGFlow\_cluster, models the mutual complex dependencies as a fully connected dynamic graph via a self-attention mechanism, so that a much more flexible relation among entities can be represented.

\subsection{Normalizing Flow for Anomaly Detection}
Normalizing flow is a powerful technique for density estimation, and it has proven successful in image generation applications~\cite{dinh2016density,papamakarios2017masked}. Recently, it has also been applied to anomaly detection tasks under the assumption that anomalies occur in low-density regions, as seen in the works of DifferNet~\cite{rudolph2021same} and CFLOW-AD~\cite{gudovskiy2022cflow}. These approaches utilize normalizing flow to estimate the likelihoods of normal embedding and identify image defects by detecting embeddings that lie far from the dense region. GANF is a great approach to employ normalizing flow for unsupervised anomaly detection in MTS data. We build on this research direction by incorporating an entity/cluster-aware normalizing flow design to increase model capacity.

\begin{figure*}
    \centering
    \includegraphics[width=1\textwidth]{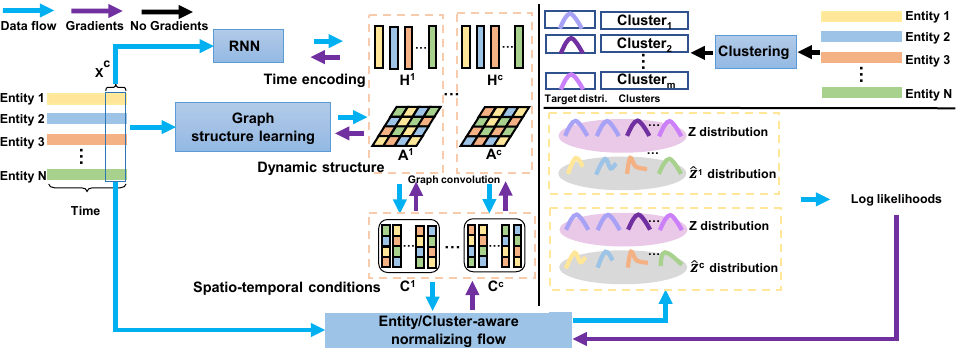}
    \caption{Overview of the proposed MTGFlow and MTGFlow\_{cluster}. 
    Within a sliding window of size $T$, time series $x^{c}$ is fed to the RNN module to capture the temporal correlations.
    Hidden states of RNN are regarded as time encoding, $H^{c}$. Meanwhile, $x^{c}$ is also input to the graph structure learning module to capture dynamic interdependencies among entities, which are modeled as adjacency matrix $A^{c}$. The spatio-temporal conditions $C^{c}$ are derived via the graph convolution operation for $H^{c}$ and $A^{c}$. Finally, $C^{c}$ is used to help entity/cluster-aware normalizing flow model to produce entity/cluster-specific density estimation for the distribution of time series.}
    \label{fig:overview}
\end{figure*}
\section{Preliminary}
We provide a brief introduction to normalizing flow in this section to better understand MTGFlow and its extension.

\subsection{Self-Attention}
Transformer~\cite{vaswani2017attention} achieves unprecedented success for natural language processing (NLP), which largely benefits from self-attention superiority in modeling relations of input sequences. Give an input sequence, $X = \left(x_1, x_2, ..., x_K\right)$, where $x_i \in \mathcal{R}^{D}$, and $K$ and $D$ are the sequence length and embedding dimension, respectively. Self-attention applies linear projections to get the query $Query \in \mathcal{R}^{K \times D}$ and key $Key \in \mathcal{R}^{K \times D}$ via learnable parameter matrices $W^{Query}\in \mathcal{R}^{D \times D}$and $W^{Key}\in \mathcal{R}^{D \times D}$. After projections, attention scores are derived by the scaled dot product between $Query$ and $Key$, and a Softmax function is utilized to normalize the scores to (0,1). 
\begin{equation}
    \centering
    \begin{split}
    Query = XW^{Query}& \quad \text{and}\quad Key = XW^{Key}\\
    Attention\_matrix = &Softmax\left(\frac{QueryKey^{T}}{\sqrt{D}}\right).
 \end{split}
 \label{equ:attention}
\end{equation}
This design allows for flexible modeling of pairwise relationships within the neural network architecture. And, the resulting attention matrix serves as the quantification of these relationships.

\subsection{Normalizing Flow}
Normalizing flow is an unsupervised technique for estimating density, achieved by mapping the original distribution to an arbitrary target distribution using a sequence of invertible affine transformations. 
When density estimation on original data distribution $\mathcal{X}$ is intractable, an alternative option is to estimate $z$ density on target distribution $\mathcal{Z}$. Let us define a source sample $x \in \mathcal{R}^{D} \sim \mathcal{X}$ and a target distribution sample $z \in \mathcal{R}^{D} \sim \mathcal{Z}$. Bijective invertible transformation $\mathcal{F}_{\theta}$ aims to achieve one-to-one mapping $z = f_{\theta}(x)$ from $\mathcal{X}$ to $\mathcal{Z}$. According to the change of variable formula, we can get 
\begin{equation}
     P_\mathcal{X}(x) = P_\mathcal{Z}(z)\left |\det\frac{\partial{f_{\theta}}}{\partial{x}^{T}}\right|.
\end{equation}
    Benefiting from  tractable jacobian determinants $\left |\det\frac{\partial{f_{\theta}}}{\partial{x}^{T}}\right|$, the objective of flow models is to achieve $\hat{z} = z$, where $\hat{z} = f_{\theta}(x)$.
    Parameters $\theta$ of ${f}_{\theta}$ can be directly estimated by MLE as follows
\begin{equation}
    \begin{split}
        \theta^{*}&=\mathop{\arg\max}\limits_{\theta}(log(P_\mathcal{X}(x)))\\
        &=\mathop{\arg\max}\limits_{\theta}(log(P_\mathcal{Z}(f_{\theta}(x)) + log(\left |\det\frac{\partial{f_{\theta}}}{\partial{x}^{T}} \right|)).
    \end{split}
\label{eq:eq7}
\end{equation}

Design of $f_{\theta}$ is an important topic in flow models. The core is to improve the flow model transferability under the premise of the tractable jacobian determinants. Representative flow model like RealNVP~\cite{dinh2016density} utilizes the affine coupling, while MAF~\cite{ papamakarios2017masked} applies autoregressive functions.

Flow models are able to achieve more superior density estimation performance when additional conditions $C$ are input~\cite{ardizzone2019guided}. This is because $C$ can usually provide relevant priors to accurately characterize the source distribution, such as time and position encoding~\cite{vaswani2017attention}. Such the flow model is called conditional normalizing flow, and its corresponding mapping is derived as $z = f_{\theta}(x|C)$. Therefore, the training objective is rewritten as 
\begin{equation}
    \theta^{*}=\mathop{\arg\max}\limits_{\theta}(log(P_\mathcal{Z}(f_{\theta}(x|C)) + log(\left |\det\frac{\partial{f_{\theta}}}{\partial{x}^{T}} \right|)).
\end{equation}

\section{Method}

\subsection{Problem Statement}
Considering a MTS dataset $\mathcal{D} =\left( x_1, x_2, ..., x_K \right)$ and $x_k \in \mathcal{R}^{L}$, where $K$ represents the total number of entities, and $L$ denotes the total number of observations of each entity. To preserve temporal correlations of the original series, we use a sliding window with size $M$ and stride size $S$ to sample the normalized MTS. $M$ and $S$ can be adjusted to obtain the training sample $x^{c}$, where c is the sampling count. $x^{c}$ is short for $x^{cS:cS+M}$. Due to the fact that anomalies exhibit unusual temporal patterns with their neighbors, introducing temporal correlation to the input will benefit detection algorithms. In this context, following~\cite{dai2021graph},  we focus on detecting window-level anomalies in this paper. That is when there exists an abnormal point in the chosen window, the window label is marked as the anomaly.
Note that the time series is normalized by z-score from different entities.
\begin{equation}
    x_k = \frac{x_k - mean(x_k)}{std(x_k)},
\end{equation}
where $mean(x_k)$ and $std(x_k)$ represent the mean and standard deviation of the k-$th$ entity along the time dimension, respectively.

\subsection{Overview of MTGFlow and MTGFlow\_{cluster}}
The main concepts behind MTGFlow and MTGFlow\_{cluster} are to dynamically model mutual dependencies so that fine-grained density estimation of the multivariate time series can be obtained.
This precise estimation allows for better detection of low-density regions, even in datasets with high anomaly contamination during training. The overview of MTGFlow and MTGFlow\_cluster can be seen in Fig.~\ref{fig:overview}. Concretely, each entity's temporal variations are modeled using an RNN, while a graph structure learning module is used to model the dynamic interdependencies. The RNN's time encoding output is then combined with the graph convolution operation, using the corresponding learned graph structure. These outputs are called spatio-temporal conditions, as they contain both temporal and structural information. In MTGFlow, these conditions are then input into the entity-aware normalizing flow module to achieve precise fine-grained density estimation. And MTGFlow\_{cluster} is designed to first cluster these entities and then estimate the density via cluster-aware normalizing flow. As we do not have the cluster prior, we utilize time series clustering methods like KShape~\cite{paparrizos2015k} to perform group assignments. Finally, the deviations of $\hat{z}$ and $z$ are measured by log likelihoods, and all modules of MTGFlow and MTGFlow\_cluster are optimized together using MLE.

\subsection{Self-Attention based Graph Structure Learning}

In order to capture the mutual dependencies that exist between entities in a multivariate time series, we employ self-attention to learn a dynamic graph structure. Each entity is treated as a graph node. Given the window sequence $x^{c}\in \mathcal{R}^{K \times M}$, we calculate the query and key vectors for each node $i$ as $x_{i}^{c}W^{Query}$ and $x_{i}^{c}W^{Key}$, where $W^{Query}\in \mathcal{R}^{M \times M}$ and $W^{Key}\in \mathcal{R}^{M \times M}$ are the query and key weights, respectively. The pairwise relationship $e_{ij}^{c}$ between node $i$ and node $j$ at the $c$-$th$ sampling count is then obtained as follows:
\begin{equation}
e_{ij}^{c} = \frac{(x_{i}^{c}W^{Query})(x_{j}^{c}W^{Key})^{T}}{\sqrt{M}}.     
\end{equation}
The attention score $a_{ij}^{c}$ is used to 
quantify the pairwise relation from node $i$ to node $j$, calculated by:
\begin{flalign}
 a_{ij}^{c}&  = \frac{\exp (e_{ij}^{c})}{\sum_{j = 1}^{K}exp({e_{ij}^{c}})}\text{,} \quad A^{c} = \begin{bmatrix}
a_{11}^{c} & \cdots & a_{1K}^{c} \\
\vdots & \ddots & \vdots \\
 a_{K1}^{c} & \cdots &  a_{KK}^{c}
\end{bmatrix}.
\end{flalign}
The resulting attention score is used to construct the attention matrix $A^c$, which serves as the adjacency matrix of the learned graph. Since input time series are evolving over time, $A^{c}$ also changes to capture the dynamic interdependencies.

\subsection{Spatio-temporal Condition}
To accurately estimate the density of multiple time series, it is crucial to incorporate robust spatio-temporal contextual information. This involves modeling the underlying structure as a dynamic graph and considering both spatial and temporal correlations to effectively capture the features of the time series. We use recurrent neural networks (RNNs) to capture temporal correlations. For a window sequence of $k$-$th$ entity, $x^{c}_k$, the time representation $H_{k}^{t}$ at time $t\in [cS:cS+M)$ is derived by $H_{k}^{t} = RNN({x_{k}^{t}, H_{k}^{t-1}})$, where RNN can be any sequence model such as LSTM~\cite{hochreiter1997long} and GRU~\cite{cho2014learning}, and $H_{k}^{t}$ is the hidden state of RNN. 

To obtain the spatio and temporal information $C^{t}$ of all entities at $t$, a graph convolution operation is performed through the learned graph $A^{c}$. As mentioned in GANF, we also find that history information of the node itself helps enhance temporal relationships of time series. Hence, the spatio-temporal condition at $t$:
\begin{equation}
    C^{t} = ReLU(A^{c}H^{t}W_{1} + H^{t-1}W_{2})W_{3},
\end{equation}
where $W_{1}$ and $W_{2}$ are graph convolution and history information weights, respectively. $W_{3}$ is used to improve the expression ability of condition representations. The spatio-temporal condition $C^{c}$ for window $c$ is the concatenation of $C^{t}$ along the time axis.

\subsection{Entity/Cluster-aware Normalizing Flow}
Distributions of individual entities have discrepancies because of their different work mechanisms, and thus their respective anomalies will generate distinct sparse characteristics.
 If we map time series from all entities to the same distribution $\mathcal{N}(0,I)$, as does in GANF, then the description capacity of the model will be largely limited, and the unique inherent property of each entity will be ignored. Therefore, we design the entity-aware normalizing flow $z_k = f^{k}_{\theta}(x|C)$ to make more detailed density estimation, where $x$, $C$, $k$ are the input sequence, condition and the $k$-$th$ entity, respectively. 
Technically, 
in MTGFlow, for one entity, we assign the multivariate Gaussian distribution as the target distribution. As for MTGFlow\_cluster, due to the absence of cluster priors, we leverage KShape for clustering all entities and subsequently assign the same target distribution to entities within a cluster. The covariance matrix of the above target distribution is the identity matrix $I$ for better convergence. Moreover, in order to generate corresponding target distributions $\mathcal{Z}_k$, mean vectors $\mu_k \in 
 R^{M}$ may be independently drawn from $\mathcal{N}(0,I)$~\cite{izmailov2020semi}. However, in our experiments, we find it better to keep each element of $\mu_k$ the same. Specifically, for the time series of the entity $k$, the entity-aware density estimation is given by:
\begin{equation}
    \begin{gathered}
    P_{\mathcal{X}_k}(x_k) = P_{\mathcal{Z}_k}(f_{\theta}^{k}(x_k|C))\left       |\det\frac{\partial{f_{\theta}^{k}}}{\partial{x_k}^{T}} \right|,\\
    \mathcal{Z}_k = \mathcal{N}(\mu_k, I), \\
    \mu_k \sim \mathcal{N}(0, 1).\\
    \end{gathered}
\end{equation}
The cluster-aware density estimation is given by:
\begin{equation}
    \begin{gathered}
    P_{\mathcal{X}_k}(x_k) = P_{\mathcal{Z}_k}(f_{\theta}^{k}(x_k|C))\left       |\det\frac{\partial{f_{\theta}^{k}}}{\partial{x_k}^{T}} \right|,\\
    \mathcal{Z}_k = \mathcal{N}(\mu_m, I) \enspace if \enspace k \in cluster_m,\\
    \mu_m \sim \mathcal{N}(0, 1),\\
       {\begin{Bmatrix} 
  cluster_1, \dots, cluster_m
\end{Bmatrix}}= KShape\left \{ \mathcal{D} \right \}.
    \end{gathered}
\end{equation}
Note that when the cardinality of all clusters is one, MTGFlow is equal to MTGFlow\_{cluster}.
In such a case, model parameters will increase with the number of entities/clusters. To mitigate this problem, we share entity/cluster-aware normalizing flow parameters across all entities. So, the density estimation for $k$ reads:
\begin{equation}
    P_{\mathcal{X}_k}(x_k) = P_{\mathcal{Z}_k}(f_{\theta}(x_k|C))\left       |\det\frac{\partial{f_{\theta}}}{\partial{x_k}^{T}} \right|.
\end{equation}

\subsection{Joint Optimization}
As described above, MTGFlow and MTGFlow\_{cluster} combine graph structure learning and RNN to capture the spatio and temporal dependencies on multiple time series. Then, derived saptio-temporal conditions are utilized to contribute to entity/cluster-aware normalizing flow estimating the density of time series. To avoid getting stuck in the local optimum for each module, we jointly optimize all modules. The whole parameters $W^*$ are estimated via MLE.
\begin{equation*}
    \begin{split}
        W^*&=\mathop{\arg\max}\limits_{W}log(P_\mathcal{X}(x)) \\
        &\approx\mathop{\arg\max}\limits_{W}\frac{1}{N}\sum^N_{c=1}\frac{1}{K}\sum^K_{k=1}log(P_{\mathcal{X}_k}(x_{k}^{c}))\\ 
        &\approx\mathop{\arg\max}\limits_{W}\frac{1}{NK}\sum^N_{c=1}\sum^K_{k=1}log(P_{\mathcal{Z}_k}(f_{\theta}(x_{k}^{c}|C_{k}^{c}))\left|\det\frac{\partial{f_{\theta}}}{\partial{x_{k}^{c}}^{T}} \right|)\\
        &\approx \mathop{\arg\max}\limits_{W}\frac{1}{NK}\sum^N_{c=1}\sum^K_{k=1}(-\frac{1}{2}\lVert \hat{z_{k}^{c}} -\mu_k\rVert_{2}^2 + log\left|\det\frac{\partial{f_{\theta}}}{\partial{x_{k}^{c}}^{T}} \right|
        \\& \quad\quad\quad\quad\quad +Const),
    \end{split}
\end{equation*}
where $N$ is the total number of windows, and $Const$ is equal to $-\frac{T}{2}*log(2\pi)$.

\subsection{Anomaly Detection and Interpretation}
Based on the hypothesis that anomalies tend to be sparse on data distributions, low log likelihoods indicate that the observations are more likely to be anomalous. 
\subsubsection{Anomaly Detection}
Taking the window sequence $x^{c}$ as the input, the density of all entities can be estimated.
The mean of the negative log likelihoods of all entities serves as the anomaly score $S_c$, which is calculated by:
\begin{equation}
    S_c =  -\frac{1}{K}\sum^K_{k=1}log(P_{\mathcal{X}_k}(x_{k}^{c})).
    \label{equ:anomalyscore}
\end{equation}
A higher anomaly score represents that $x_{k}^{c}$ is located in the lower density region, indicating a higher possibility to be abnormal. Since abnormal series exist in the training set and validation set, we cannot directly set the threshold to label the anomaly, such as the maximum deviation in validation data~\cite{deng2021graph}. Therefore, to reduce the anomaly disturbance, we store $S_c$ of the whole training set, and the interquartile range (IQR)  is used to set the threshold:
\begin{equation}
    Threshold = Q_3 + 1.5*(Q_3 - Q_1),
\end{equation}
where $Q_1$ and $Q_3$ are 25-$th$ and 75-$th$ percentile of $S_{c}$. 
\subsubsection{Anomaly Interpretation}
Abnormal behaviors of any entity could lead to the overall abnormal behavior of the whole window sequence.
Naturally, we can get the anomaly score $S_{ck}$ for entity $k$ according to Eq.~\eqref{equ:anomalyscore}.
\begin{equation}
    S_c =  -\frac{1}{k}\sum^K_{k=1}log(P_{\mathcal{X}_k}(x_{k}^{c})) = \sum^K_{k=1}S_{ck}.
\end{equation}

Since we map the time series of each entity into unique target distributions, different ranges of $S_{ck}$ are observed. This bias will assign each entity to different weights in terms of its contribution to $S_c$. To circumvent the above-unexpected bias, we design the entity-specific threshold for each entity. 
Considering different scales of $S_{ck}$, IQR is used to set respective thresholds. Therefore, the threshold for $S_{ck}$ is given as: 
\begin{equation}
    Threshold_{k} = \lambda_k(Q_3^k + 1.5*(Q_3^k - Q_1^k)),
    \label{equ:entityanomalyscore}
\end{equation}
where $Q_1^k$ and $Q_3^k$ are 25-$th$ and 75-$th$ percentile of $S_{ck}$ across all observations, respectively. And $\lambda_k $ is used to adjust $Threshold_{k}$ because normal observations in different entities also fluctuate with different scales.

\begin{table*}[]
    \caption{The static and settings of five public datasets.}
    \small
    \centering
    \begin{tabular}{c|c|c|c|c|c}
    \toprule
      Dataset & Entity / metric number & Training set & Training set anomaly ratio (\%)& Testing set & Testing set anomaly ratio (\%)\\
        \midrule
      \rule{0pt}{10pt} SWaT &  51 & 269951 & 17.7& 89984 & 5.2\\
      \rule{0pt}{10pt} WADI & 123 &  103680 &  6.4 & 69121 &4.6\\
      \rule{0pt}{10pt} PSM & 25 &  52704 &  23.1 & 35137 &34.6\\ 
      \rule{0pt}{10pt} MSL & 55 &  44237 &  14.7 & 29492 &4.3\\ 
      \rule{0pt}{10pt} SMD & 38 &  425052 &  4.2 & 283368 &4.1\\ 
      \bottomrule
    \end{tabular}

    \label{tab:dataset_setting}
\end{table*}

\section{Experiment}
\label{Experiment}
\subsection{Experiment Setup}
\subsubsection{Dataset}
The commonly used public datasets for MTS anomaly detection in OCC are as follows:
\begin{itemize}
     
\item SWaT~\cite{goh2016dataset} is a collection of 51 sensor data from a real-world industrial water treatment plant, at the frequency of one second. The dataset provides ground truths of 41 attacks launched during 4 days.

\item WADI~\cite{ahmed2017wadi} is a collection of 123 sensor and actuator data from WADI testbed, at the frequency of one second. The dataset provides ground truths of 15 attacks launched over 2 days.

\item PSM~\cite{abdulaal2021practical} is a collection of multiple application server nodes at eBay with 25 features. The datasets provide ground truths created by experts over 8 weeks.

\item MSL~\cite{hundman2018detecting} is a collection of the sensor and actuator data of the Mars rover with 55 dimensions.

\item SMD~\cite{su2019robust} is comprised of 28 small datasets from 28 machines at an internet company that records server metric data like CPU utilization, at the frequency of one minute. The dataset provides labeled anomalies for 5 weeks.
\end{itemize}

Since only normal time series are provided in these datasets for training in OCC setting, we follow the dataset setting of GANF~\cite{dai2021graph} and split the original testing dataset by 60\% for training, 20\% for validation, and 20\% for testing in SWaT. For other datasets, the training split contains 60\% data, and the test split contains 40\% data. Dataset statistics are provided in Table~\ref{tab:dataset_setting}. 

In addition, we also consider the common OCC setting for a comprehensive evaluation. Under the OCC setting, the training datasets are required to be normal data, so we use the original training datasets of the above-mentioned datasets as training datasets.
Besides, Wu~\emph{et al.}~\cite{wu2021current} provides elaborate and challenging datasets (UCR) including 250-sub univariate datasets from diverse real scenarios (e.g., from biology and industry). For every sub-dataset, the training data is free of anomalies and the test data has only one short-period anomaly. Therefore, we follow the origin split for evaluation on UCR. Although MTGFlow is designed to detect multivariate time series anomaly detection, it can also be applied to univariate series. Note that MTGFlow is equivalent to MTGFlow\_cluster when detecting univariate time series.

\begin{table*}
     \caption{Anomaly detection performance of AUROC(\%) on five public datasets. We test MTGFlow and MTGFlow\_cluster in unsupervised and OCC settings. The best results are highlighted in bold.}
    \centering
    \resizebox{\textwidth}{!}{
    \begin{tabular}{c|c|ccccccccc}

    \toprule
        \multicolumn{2}{c|}{Dataset} &DeepSVDD &ALOCC &DROCC &DeepSAD & USAD & DAGMM &GANF &MTGFlow & MTGFlow\_{cluster} \\
        \midrule
       \rule{0pt}{10pt} \multirow{5}{*}{\makecell{Unsupervised \\ setting}}& SWaT &66.8\textpm{2.0} &77.1\textpm{2.3} &72.6\textpm{3.8} &75.4\textpm{2.4} &78.8\textpm{1.0}&72.8
\textpm{3.0} &79.8\textpm{0.7} & \textbf{84.8\textpm{1.5}}  & 83.1\textpm{1.3}\\
       \rule{0pt}{10pt} & WADI  & 83.5\textpm{1.6} &83.3\textpm{1.8} &75.6\textpm{1.6}& 85.4\textpm{2.7}  &86.1\textpm{0.9} &77.2\textpm{0.9}
& 90.3\textpm{1.0} &  \textbf{91.9\textpm{1.1}}  & 91.8\textpm{0.4}\\ 
       \rule{0pt}{10pt} & PSM  & 67.5\textpm{1.4} &71.8\textpm{1.3} & 74.3\textpm{2.0}& 73.2\textpm{3.3} &78.0\textpm{0.2}  &64.6 \textpm{2.6} 
& 81.8\textpm{1.5} &  85.7\textpm{1.5}  &\textbf{87.1\textpm{2.4}}	\\
       \rule{0pt}{10pt} & MSL  & 60.8\textpm{0.4} &60.3\textpm{0.9} &53.4\textpm{1.6}& 61.6\textpm{0.6} &57.0\textpm{0.1}  &56.5 \textpm{2.6} 
& 64.5\textpm{1.9} &  67.2\textpm{1.7}  & \textbf{68.2\textpm{2.6}}\\ 
       \rule{0pt}{10pt} & SMD  & 75.5\textpm{15.5} &80.5\textpm{11.1}& 76.7\textpm{8.7}  &85.9 \textpm{11.1} &86.9\textpm{11.7} &78.0\textpm{9.2}
& 89.2\textpm{7.8} &  \textbf{91.3\textpm{7.6}}  & -\\ \hline

       \rule{0pt}{10pt} \multirow{5}{*}{\makecell{OCC \\ setting}}& SWaT &\textbf{85.9\textpm{2.6}} &78.2\textpm{2.3} &83.5\textpm{1.0} &83.7\textpm{2.0} &77.1\textpm{0.8}&73.4\textpm{0.2} &80.7\textpm{0.8} & 83.9\textpm{0.6} & 83.5\textpm{1.4}\\
       \rule{0pt}{10pt} & WADI  & 85.5\textpm{0.4} &88.7\textpm{0.9} &89.0\textpm{1.1}& 88.2\textpm{1.5} &88.5\textpm{1.2} &84.8\textpm{2.3}
& 91.7\textpm{0.9} &   \textbf{92.2\textpm{1.0}} & 92.1\textpm{0.8}\\ 
       \rule{0pt}{10pt} & PSM  & 85.5\textpm{1.9} &74.7\textpm{1.7} & 83.4\textpm{1.1}  & 84.9\textpm{1.0} &80.0\textpm{0.8}  &69.4\textpm{1.6}  
& 85.2\textpm{0.9}&  86.1\textpm{1.6} & \textbf{87.9\textpm{2.0}}\\
       \rule{0pt}{10pt} & MSL  & 61.9\textpm{0.8} &61.8
\textpm{0.9} &56.8\textpm{0.6}& 62.8\textpm{0.7} &58.9\textpm{0.6}  &59.3\textpm{0.6} 
& 66.7\textpm{1.4}  &  68.3\textpm{1.3} &\textbf{68.9\textpm{1.6}} \\ 
       \rule{0pt}{10pt} & SMD  & 86.4\textpm{9.8} &81.9\textpm{9.4}& 79.1\textpm{9.7}  &89.4\textpm{8.7} &87.8\textpm{11.3} &79.1\textpm{9.2}
& 90.2\textpm{8.4} &  \textbf{91.8\textpm{6.4}} & -\\ 
       \bottomrule	
    \end{tabular}}
    \label{tab:my_label}
\end{table*}

\subsubsection{Implementation Details}
For datasets of MTS, we set the window size as 60 and the stride size as 10. As for univariate dataset UCR, the duration of the anomalies is very short, and in some cases only a dozen sampling points. Hence, the window size is set as 10 for UCR. Adam optimizer with a learning rate of 0.002 is utilized to update all parameters. One layer of LSTM is employed to extract time representations in our experiment.
One self-attention layer with 0.2 dropout ratio is adopted to learn the graph structure.
We use MAF as the normalizing flow model. In MTGFlow\_{cluster}, KShape is used to perform the whole time series clustering. The preset number of clusters is 20.
For SWaT, one flow block and 512 batch size are employed.
For other datasets, we arrange two flow blocks for it and set the batch size as 256. $\lambda$ is set as 0.8 for thresholds of all entities. The epoch is 40 for all experiments, which are performed in PyTorch-1.7.1 with a single NVIDIA RTX 3090 24GB GPU\footnote{ Code is available at github.com/zqhang/MTGFLOW.}.

\subsubsection{Evaluation Metric}
As in previous works, MTGFlow and MTGFlow\_cluster aim to detect the window-level anomalies, and labels are annotated as abnormal when there exists any anomalous time point. The performance is evaluated by the Area Under the Receiver Operating Characteristic curve (AUROC).

\subsubsection{Baselines}
To illustrate the effectiveness of our proposed method, we compare the SOTA semi-supervised and unsupervised density estimation methods.
\begin{itemize}
\item DeepSAD~\cite{ruff2019deep}. A semi-supervised method maps normal data into a preset hypersphere, and additional anomaly labels are leveraged to improve the detection performance.
\item DeepSVDD~\cite{ruff2018deep}. An OCC-based method maps normal data into a preset hypersphere, and anomaly data are assumed to be located outside the hypersphere.
\item ALOCC~\cite{sabokrou2020deep}. An OCC-based method reconstructs inputs via GAN and declares anomalies according to the output of the discriminator.
\item DROCC~\cite{goyal2020drocc}. An OCC-based method constructs robust representations for normal data. Anomalies will deviate from the learned representations.
\item USAD~\cite{audibert2020usad}. An OCC-based method leverages Autoencoder to reconstruct inputs. Anomaly data suffer poor reconstruction quality. 
\item DAGMM~\cite{zong2018deep}. A density estimation approach combines Autoencoder and Gaussian Mixture Model. Samples with high energy are declared as anomalies.
\item GANF~\cite{dai2021graph}. The density estimation approach joints DAG and flow model, and data on low-density regions are identified as anomalies.
\end{itemize}

\subsubsection{Performance}
\label{Performance}

\begin{figure*}
  \centering
    \subfloat[Roc curves of SWaT ]{\includegraphics[width=0.25\textwidth]{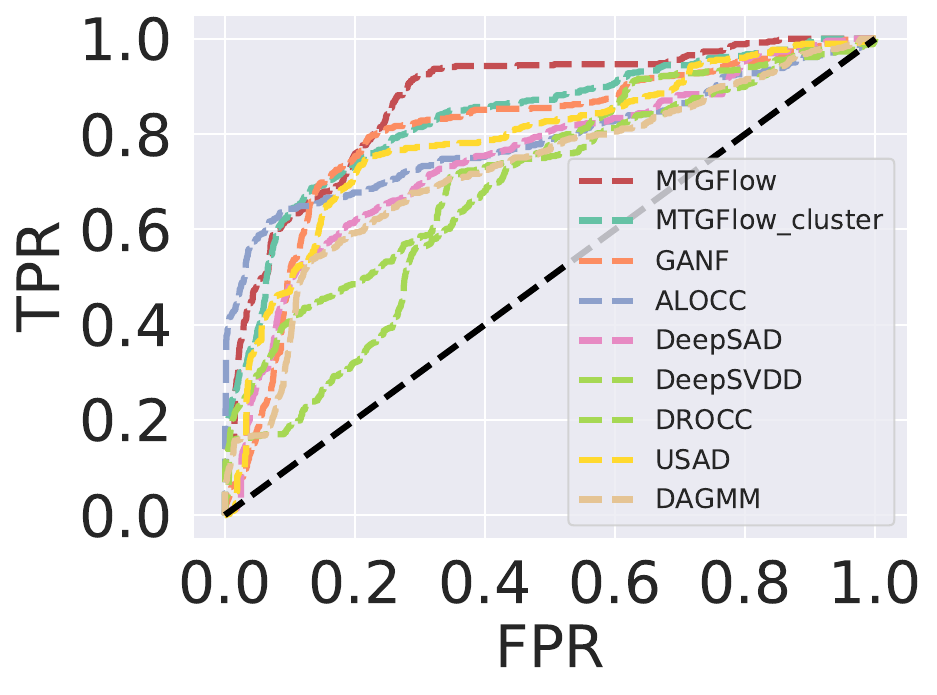}%
    \label{fig: SWAT_ROC}}
    \hfil
    \subfloat[Roc curves of WADI]{\includegraphics[width=0.25\textwidth]{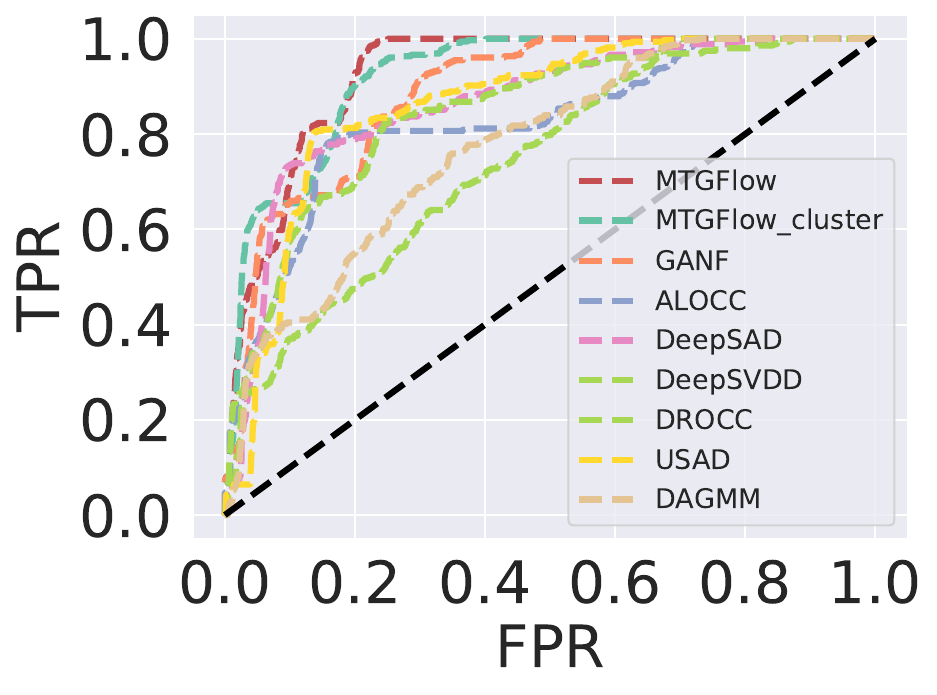}%
    \label{fig: WADI_ROC}}
    \hfil
    \subfloat[Roc curves of PSM]{\includegraphics[width=0.25\textwidth]{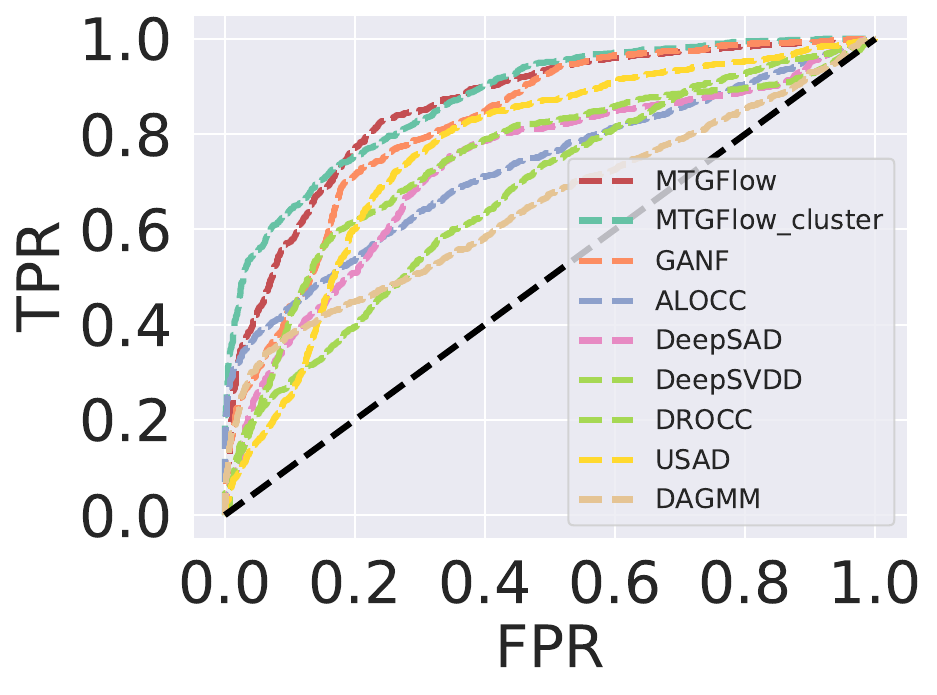}%
    \label{fig: PSM_ROC}}
    \hfil
    \subfloat[Roc curves of MSL]{\includegraphics[width=0.25\textwidth]{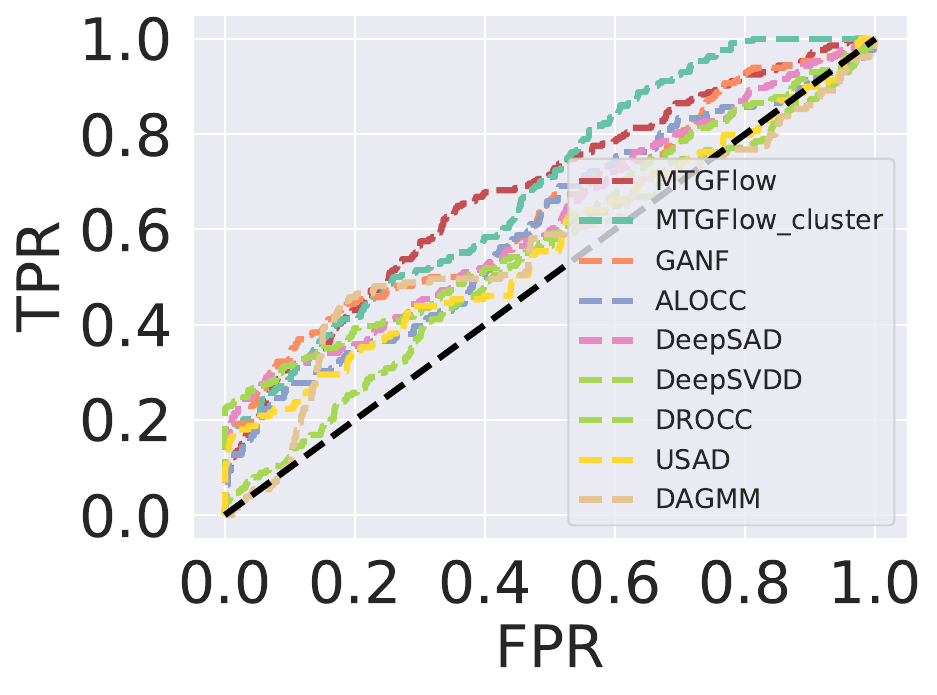}%
    \label{fig: MSL_ROC}}
 \caption{Comparison with advanced methods on ROC curves for four datasets.}
 \vspace{-1em}
 \label{fig:auroc_curves}
\end{figure*}

\begin{figure}
    \centering
    \includegraphics[width=1\columnwidth]{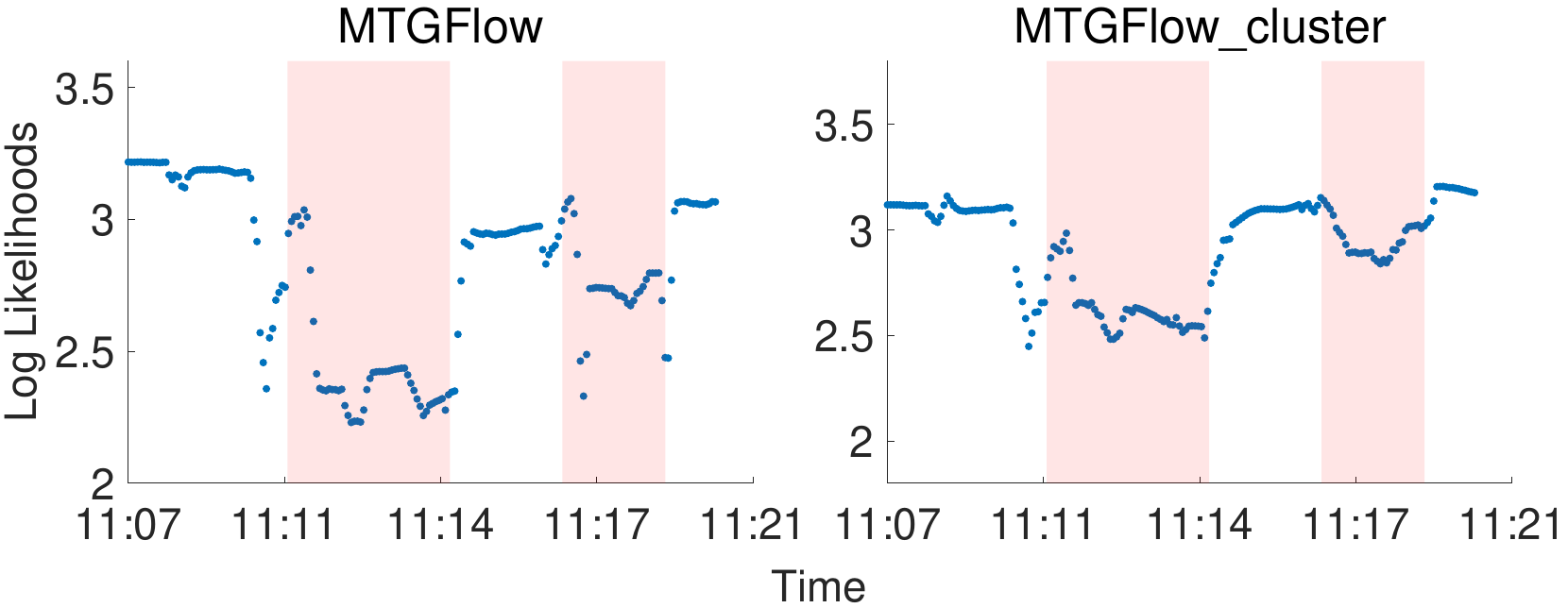}
    \caption{Log likelihoods for anomalies on MTGFlow and MTGFlow\_cluster.}
    \label{fig: Log likelihoods for anomalies}
\end{figure}

\begin{figure}
    \centering
    \includegraphics[width=1\columnwidth]{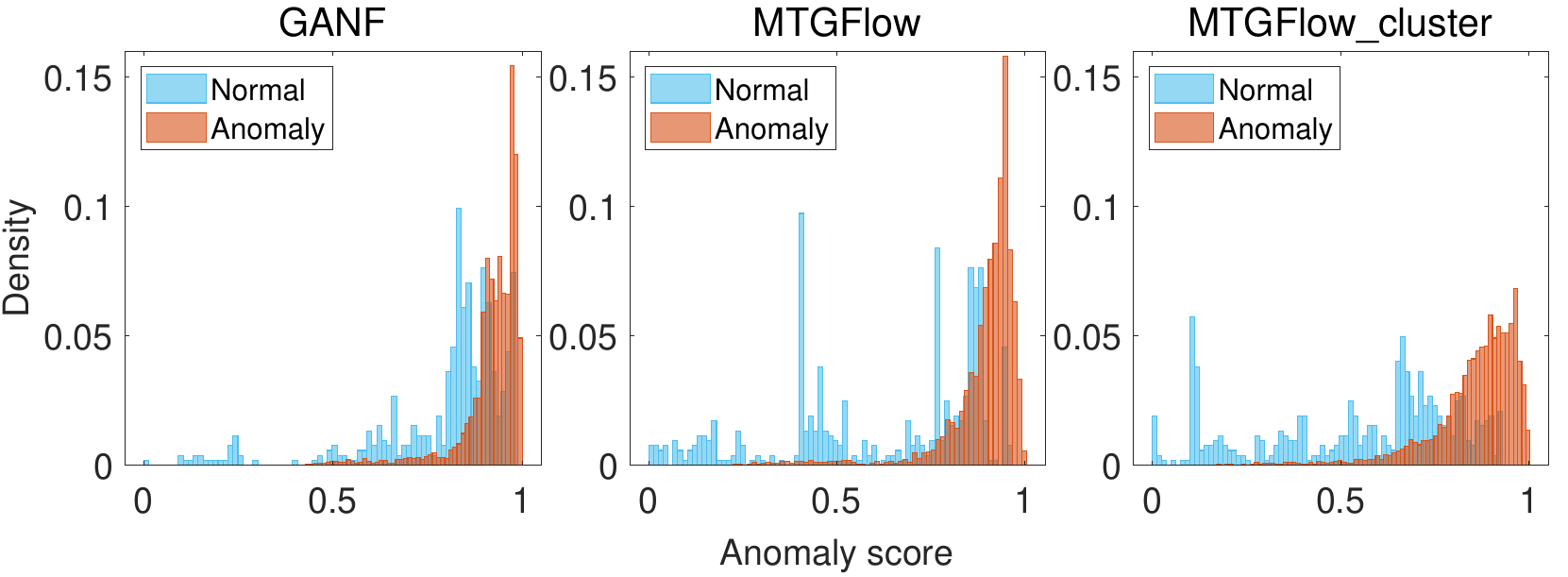}
    \caption{Discrimination comparison on SWaT between GANF, MTGFlow, and MTGFlow\_cluster.}
    \label{fig: comparison on anomaly scores between GANF and MTGFlow}
\end{figure}

\begin{table}
    \caption{Performance of AUROC(\%) on UCR.}
 
    \centering
    \begin{tabular}{c|cccc}
    \toprule
         \multirow{4}{*}{\makecell{Dataset \\ \\ UCR}}&DeepSVDD&ALOCC&DROCC&DeepSAD \\ &52.1\textpm{1.3}&58.6\textpm{1.9}&59.3\textpm{2.4}&60.5\textpm{2.1} \\  \cline{2-5}
         &USAD&DAGMM&GANF&MTGFlow \\ 
        &56.7\textpm{0.0}&62.5\textpm{0.5}&63.4\textpm{0.2}&\textbf{67.4\textpm{0.4}}  \\
 
       \bottomrule	
    \end{tabular}
  
    \label{tab:ucr_table}
\end{table}

\begin{figure*}
    \centering
    \includegraphics[width=1\textwidth]{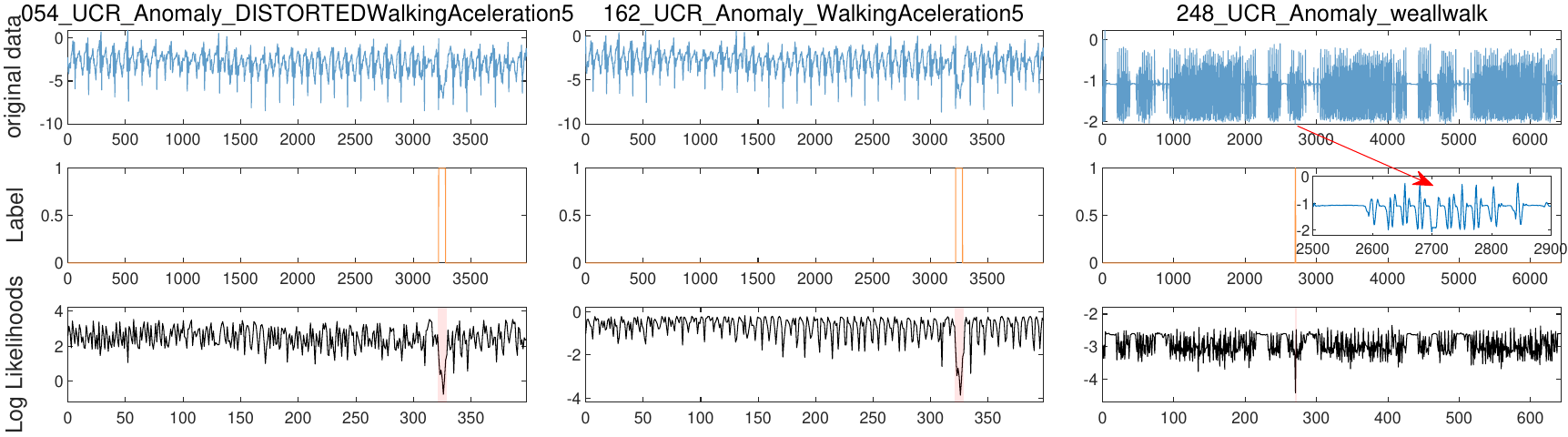}
    \caption{Log likelihoods for anomalies on UCR. \textbf{Top}: The original test data in respective sub-datasets are presented. \textbf{Middle}: The corresponding labels provided in these sub-datasets are provided. Normal time points are labeled as 0, while abnormal ones are labeled as 1. \textbf{Bottom}: Log likelihoods are obtained from our method, and the ground truths of abnormal points are highlighted in red.}
    \label{fig: log_time_ucr}
\end{figure*}

\paragraph{Detection for multivariate time series}
We list the AUROC metric results in Table~\ref{tab:my_label}, which is the average of five runs, in addition to the standard variance. Note that we do not provide the AUROC curves of SMD because it consists of 28 small datasets, where we test the performance on each of them and average all the results. MTGFlow and MTGFlow\_cluster have superior performance over all the other seven baselines under the unsupervised setting. To present an intuitive result, AUROC curves are also presented in Fig.~\ref{fig:auroc_curves}. Compared with MTGFlow and MTGFlow\_cluster, DeepSVDD and DROCC project all training samples into the hypersphere so they cannot learn the accurate decision boundary distinguishing normal from abnormal samples. Adversarial learning used by ALOCC and USAD and the semi-supervised learning strategy in DeepSAD leverages a more informative training procedure to mitigate the effect of high anomaly contamination. As for DAGMM, it is restricted to the distribution estimation ability of GMM for multiple entities. Although GANF obtains a better result, its detection performance is still limited by inadequate dependence modeling and indiscriminative density estimation. 
Due to a much more flexible modeling structure, MTGFlow outperforms the above baseline methods. Moreover, we study log likelihoods for anomalies ranging from 2016/1/2 11:07:00 to 11:37:00 in Fig.~\ref{fig: Log likelihoods for anomalies}. It is clear that log likelihoods are high for the normal series but lower for labeled abnormal ones (highlighted in red). This variation of log likelihoods validates that MTGFlow and MTGFlow\_cluster can detect anomalies according to low-density regions of modeled distribution. Meanwhile, to investigate the anomaly discrimination ability of MTGFlow and MTGFlow\_cluster, we present the normalized $S_c$ for GANF, MTGFlow, and MTGFlow\_cluster in Fig.~\ref{fig: comparison on anomaly scores between GANF and MTGFlow}. As it is displayed, for normal series, anomaly scores of MTGFlow and MTGFlow\_cluster are more closed at 0 than those of GANF, and the overlap areas of normal and abnormal scores are also smaller, reducing the false positive ratio. The larger score discrepancy corroborates MTGFlow and MTGFlow\_cluster superior detection performance.

For the OCC setting, MTGFlow and MTGFlow\_cluster also achieve promising performance.
Interestingly, the performance of MTGFlow and MTGFlow\_cluster are close to that in the unsupervised setting, while other baselines have a tremendous performance improvement on different datasets. Specifically, DeepSVDD has a significant gain from 66.8 AUROC\% to 85.9 AUROC\% on SWaT and from 67.5 AUROC\% to 85.5 AUROC\% on PSM. Besides, DROCC also obtains the advancement from 75.6 AUROC\% to 89.0 AUROC\% on WADI. These observations reflect that the anomaly contamination severely degrades the detection performance of the above OCC-based methods. Therefore, such contrast illustrates that our method is more robust to anomaly contamination and thus applicable to both two scenarios. 

    Although MTGFlow and MTGFlow\_cluster are superior to the baselines, we observe that MTGFlow\_cluster does not consistently outperform MTGFlow according to the results. We attribute this performance degradation (in the case of SWaT and WADI datasets) to the inappropriate cluster assignments. Because of the lack of prior clusters, we use KShape to provide the pseudo cluster label. Although KShape is a highly accurate and scalable time series clustering algorithm, the entity allocation is sensitive to the number of predefined clusters. If two or more entities with distinct characteristics are clustered into a group, MTGFlow\_cluster will combine multiple underlying data manifolds into a single distribution, thereby hindering the accurate estimation for each entity. In our experiments, we apply default parameters to all datasets, i.e., the number of predefined clusters is set to 20 across different datasets such as SWaT and WADI. This cluster number may be not optimal for all datasets. To investigate the effect of cluster number, we perform a detailed analysis in Section~\ref{sec: cluster}.

\paragraph{Detection for univariate time series}
As for univariate series, we evaluate our method on UCR with other baselines. Since UCR has 250 sub-datasets, we first average the results over these sub-datasets and then average the above results over five runs with a standard variance.  From Table~\ref{tab:ucr_table}, our method still obtains the best performance. As suggested in~\cite{wu2021current},
We visual the original data of three sub-datasets (i.e., the unexpected acceleration of distorted walking and walking)
and present the obtained log likelihoods from our method in Fig.~\ref{fig: log_time_ucr}. From the figure, anomalous points in $054\_UCR$ and $162\_UCR$ datasets are not sudden extreme values, which may be identified via an empirical threshold. Their values are in the normal range but exhibit an unusual pattern caused by the unexpected acceleration. Furthermore, in addition to the difficulty mentioned above, the duration of the anomaly in $248\_UCR$ dataset is also very short (i.e., just five abnormal points). These challenges make it more difficult to detect such anomalies. Bypassing these problems, our method estimates the density of data points rather than focusing on their numerical values and successfully detects such anomalies according to the figure. Moreover, we can observe that anomalous points incur lower log likelihoods than normal points. And, the prominent peaks show that MTGFlow can discriminate anomalies clearly.

\begin{table}
    \caption{Module ablation study (AUROC\%).}
        \small
    \centering
    \footnotesize
    \begin{tabular}{c|c|c|c|c}
    \toprule
         &Graph & Entity-aware &SWaT &WADI  \\
        \midrule
       \rule{0pt}{10pt} MTGFlow/(G, E)   &\XSolidBrush &\XSolidBrush & 78.3\textpm{0.9} & 89.7\textpm{0.5} \\
       \rule{0pt}{10pt} MTGFlow/G   &\XSolidBrush &\Checkmark & 82.4\textpm{1.0} & 91.3\textpm{0.4}  \\
       \rule{0pt}{10pt} MTGFlow/E  &\Checkmark&\XSolidBrush  & 81.2\textpm{1.1} & 91.0\textpm{0.7} \\
       \rule{0pt}{10pt} MTGFlow     &\Checkmark& \Checkmark& \textbf{84.8\textpm{1.5}} & \textbf{91.9\textpm{1.1}}  \\
    \bottomrule
    \end{tabular}

    \label{tab:ablation_module design}
\end{table}

\begin{table}
    \caption{Ablation study of hyperparameters (AUROC\%).}
    \small
    \centering
    \begin{tabular}{c|c|c|c|c}
    \toprule
        \multicolumn{2}{c|}{\diagbox[width=10em]{Window size}{Blocks}}&1&2&3  \\\midrule
        \multirow{5}{*}{SWaT}\rule{0pt}{10pt}&40&81.4\textpm{3.2}&82.7\textpm{2.1} &81.7\textpm{0.9}\\\cline{2-5}
        \rule{0pt}{10pt}&60&\textbf{84.8\textpm{1.5}} &83.6\textpm{2.0}&83.1\textpm{0.9}\\\cline{2-5}
        \rule{0pt}{10pt}&80&82.8\textpm{1.0}&82.7\textpm{0.8}&83.4\textpm{0.6}\\\cline{2-5}
        \rule{0pt}{10pt}&100&82.6\textpm{0.5}&83.4\textpm{0.9}&83.5\textpm{0.6}\\\cline{2-5}
        \rule{0pt}{10pt}&120&83.2\textpm{2.0} &83.4\textpm{2.3}&84.5\textpm{2.6}\\\cline{1-5}
        \multirow{5}{*}{WADI}\rule{0pt}{10pt}&40&90.8\textpm{1.3}&91.7\textpm{1.2} &91.7\textpm{1.3}\\\cline{2-5}
        \rule{0pt}{10pt}&60&89.2\textpm{1.9} &\textbf{91.9\textpm{1.1}}&91.5\textpm{0.8}\\\cline{2-5}
        \rule{0pt}{10pt}&80&89.8\textpm{2.0}&90.7\textpm{0.8}&91.7\textpm{0.7}\\\cline{2-5}
        \rule{0pt}{10pt}&100&89.6\textpm{1.1}&90.9\textpm{0.8}&91.8\textpm{0.6}\\\cline{2-5}
        \rule{0pt}{10pt}&120&88.6\textpm{1.4} &91.0\textpm{0.6}&91.5\textpm{0.9}\\
     \bottomrule
    \end{tabular}
    \label{tab:ablation_hyperparameters}
\end{table}

\begin{figure}[t]
    \centering
    \includegraphics[width=0.9\columnwidth]{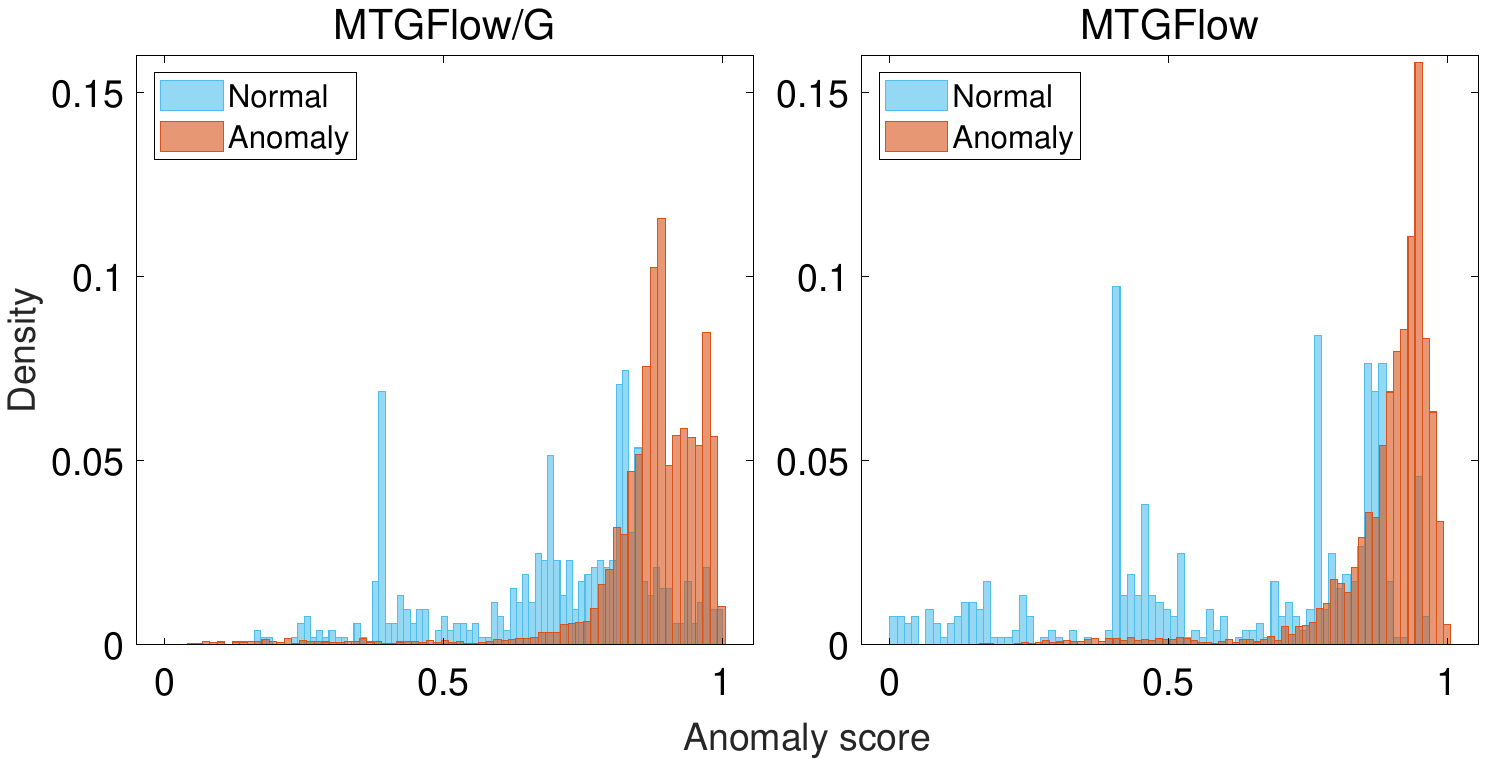}
    \caption{Comparison on normalized anomaly scores between MTGFlow and MTGFlow/G.}
    \label{fig:noattention}
\end{figure}

\begin{figure}[t]
    \centering
    \includegraphics[width=0.9\columnwidth]{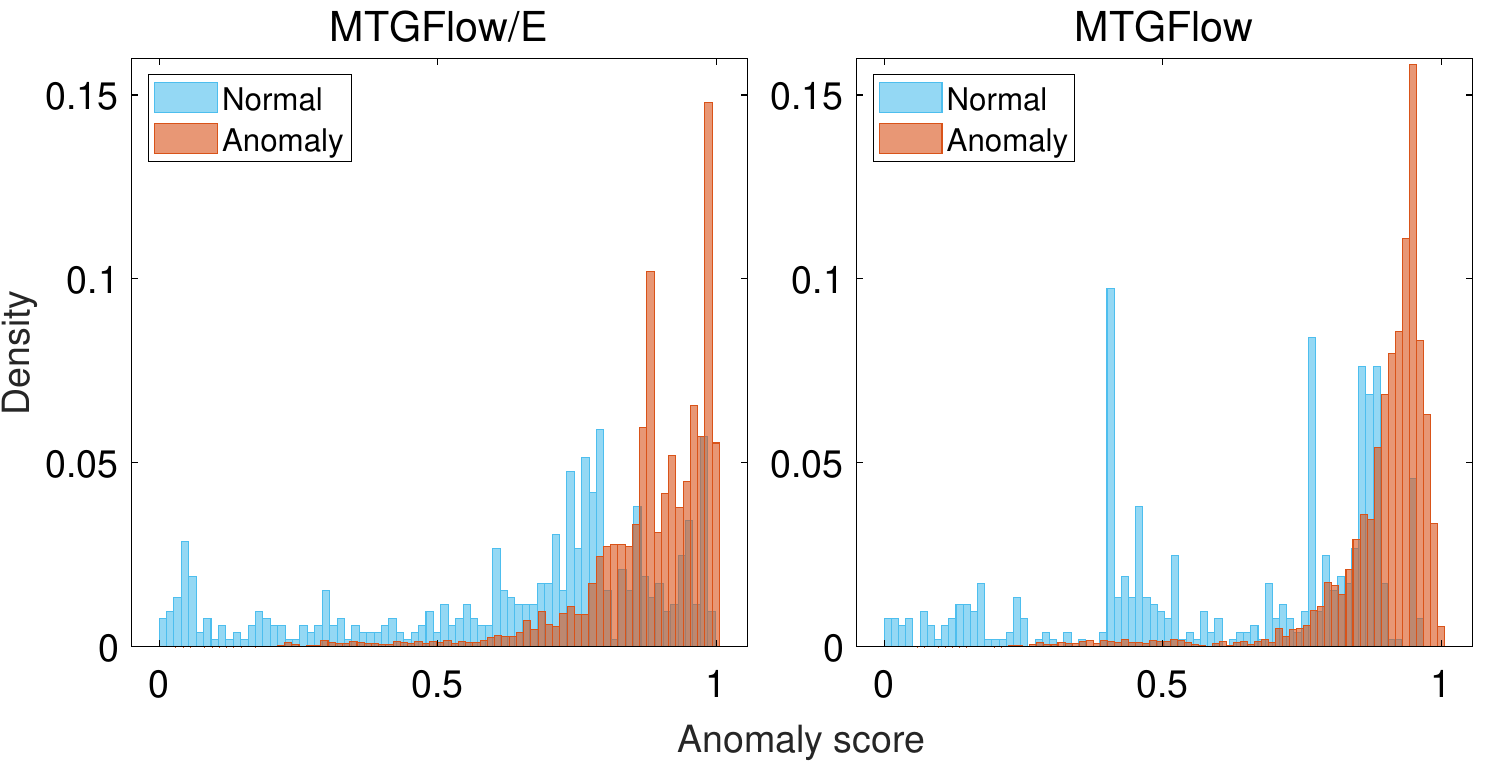}
    \caption{Comparison on normalized anomaly scores between MTGFlow and MTGFlow/E.}
    \label{fig: zero}
\end{figure}


\subsection{Ablation Study}
\label{Ablation Study}
To study the effectiveness of our proposed module, we perform several ablation studies, including module ablation, hyperparameter robustness, and anomaly ratio analysis. We mainly focus on the analysis of MTGFlow because it is the special case of MTGFlow\_{cluster}, where the cardinality of each cluster is one. As for the effect of cluster quantities, we will analyze it in Sec~\ref{sec: cluster}.
\subsubsection{Module Ablation Study}
To test the validity of each designed module, we give several ablation experiments.
We denote MTGFlow without graph and entity-aware normalizing flow as MTGFlow/(G, E), MTGFlow only without graph as MTGFlow/G, and MTGFlow only without entity-aware normalizing flow as MTGFlow/E. Results are presented in Table~\ref{tab:ablation_module design}, where MTGFlow/(G, E) obtains the worst performance. It is attributed to the lack of relational modeling among entities and
indistinguishable density estimation. Applying graph structure learning to model pairwise relations, MTGFlow/E achieves better performance. Also, considering more fine-grained density estimation, MTGFlow/G achieves an improvement over MTGFlow/(G, E). Integrating these two modules, MTGFlow accomplishes the best results. To provide a more intuitive result, we visualize the normalized anomaly score distributions of MTGFlow and MTGFlow/G in Fig.~\ref{fig:noattention}. Such a dynamic graph design can push anomaly scores of normal series to 0 and enlarge the margin between normal and abnormal series. Also,  we present the normalized anomaly score distribution for normal and abnormal series in Fig.~\ref{fig: zero}. We observe that the margin between normal and abnormal samples in MTGFlow is larger than that in MTGFlow/E. This demonstrates that MTGFlow amplifies the discrepancy between normal and abnormal samples with the help of entity-specific density estimation.

\subsubsection{Clustering Quantity Study}
\label{sec: cluster}
\begin{figure}
    \centering
    \includegraphics[width=1\columnwidth]{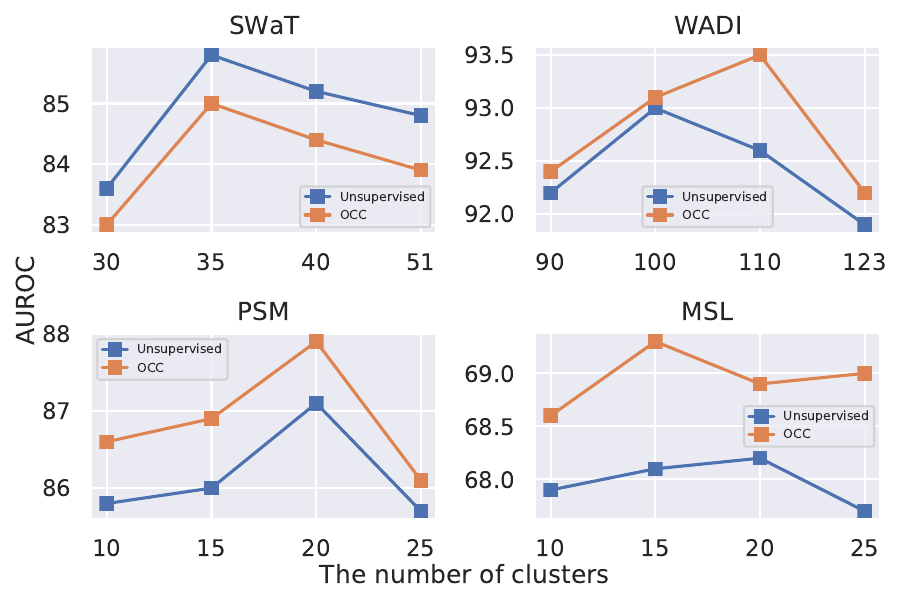}
    \caption{Effect of the cluster number in MTGFlow\_{cluster}.}
    \label{fig: cluster_num}
\end{figure}

The pre-set cluster number is important to the performance of MTGFlow\_cluster. Hence, we investigate the effect of the number of clusters on detection performance under unsupervised and OCC settings. Since the number of entities varies from different datasets, we assign different cluster numbers for them, as shown in Fig.~\ref{fig: cluster_num}. Take SWaT for example, the detection performance reaches the peak when the cluster number is 35. This illustrates that the appropriate number of clusters can facilitate the detection performance for both settings. If the number of clusters is unexpectedly small, some entities with completely different characteristics are gathered into the same cluster and transformed into the same target distribution. In this case, the detection performance will decrease, as presented by the cluster number from 35 to 30 in SWaT. Conversely, when the number of clusters is excessive, too many unnecessary target distributions need to be estimated, which may also result in performance degradation, as shown by the cluster number from 35 to 51 in SWaT. Similar phenomena are also observed across other datasets. We refer to MTGFlow\_cluster as MTGFlow\_cluster$^\ast$ when the appropriate cluster number is assigned. As shown in Table~\ref{tab: cluster_num}, MTGFlow\_cluster$^\ast$ achieves better detection performance than MTGFlow.

\begin{table}
     \caption{Performance comparison on MTGFlow and MTGFlow\_cluster$^\ast$.}
    \centering
    \resizebox{\columnwidth}{!}{
    \begin{tabular}{c|c|cccc}

    \toprule
        \multicolumn{2}{c|}{Methods} &SWaT &WADI &PSM &MSL \\
        \midrule
       \multirow{2}{*}{\makecell{Unsupervised \\ setting}}&  MTGFlow\_{cluster}$^\ast$ &85.8 &93.0&87.1&68.2
\\
   & MTGFlow&84.8 &91.9&85.7 &67.2
\\ \hline
   
       \multirow{2}{*}{\makecell{OCC \\ setting}}&  MTGFlow\_{cluster}$^\ast$ &85.0&93.5&87.9 &69.3
\\
     & MTGFlow&83.9 &92.2 &86.1 &68.3
\\

       \bottomrule	
    \end{tabular}
    }
    \label{tab: cluster_num}
\end{table}

\subsubsection{Hyperparameter Robustness}
We conduct a comprehensive study on the choice of hyperparameters, the results are shown in Table~\ref{tab:ablation_hyperparameters}. Concretely, we conduct experiments with various sizes for the sliding window and the number of the normalizing flow blocks in Table~\ref{tab:ablation_hyperparameters}. When the window size is small, such as 40, 60, and 80, the increase in the number of blocks does not necessarily improve anomaly detection performance. A larger model may cause overfitting to the whole distribution, where abnormal sequences are undesirably located in high-density regions of this distribution. When the window size is large (i.e., 80, 100, and 120), the distribution to be estimated is more high-dimensional so that model needs more capacity. Therefore, detection performance derives the average gain with blocks increasing due to more accurate distribution modeling. 

\begin{figure}[t]
    \centering
    \includegraphics[width=0.9\columnwidth]{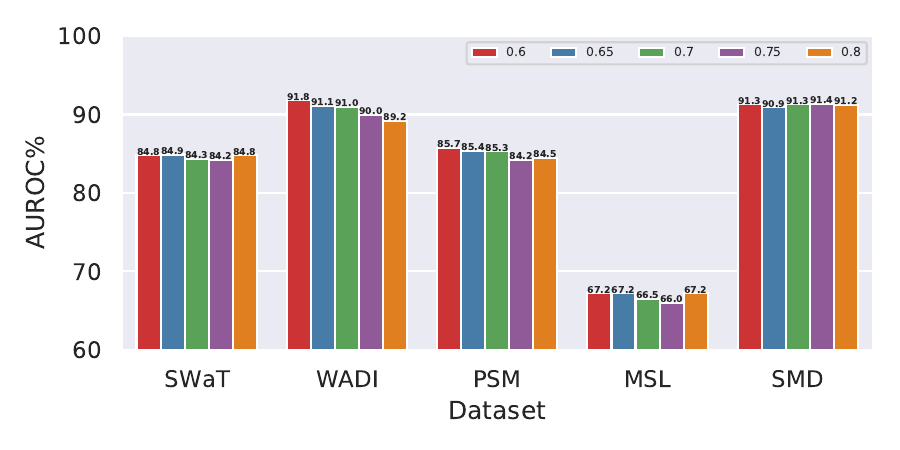}
    \caption{Effect of anomaly contamination ratio.}
    \label{fig: anomaly_ratio}
\end{figure}

\begin{figure}[t]
    \centering
    \includegraphics[width=0.9\columnwidth]{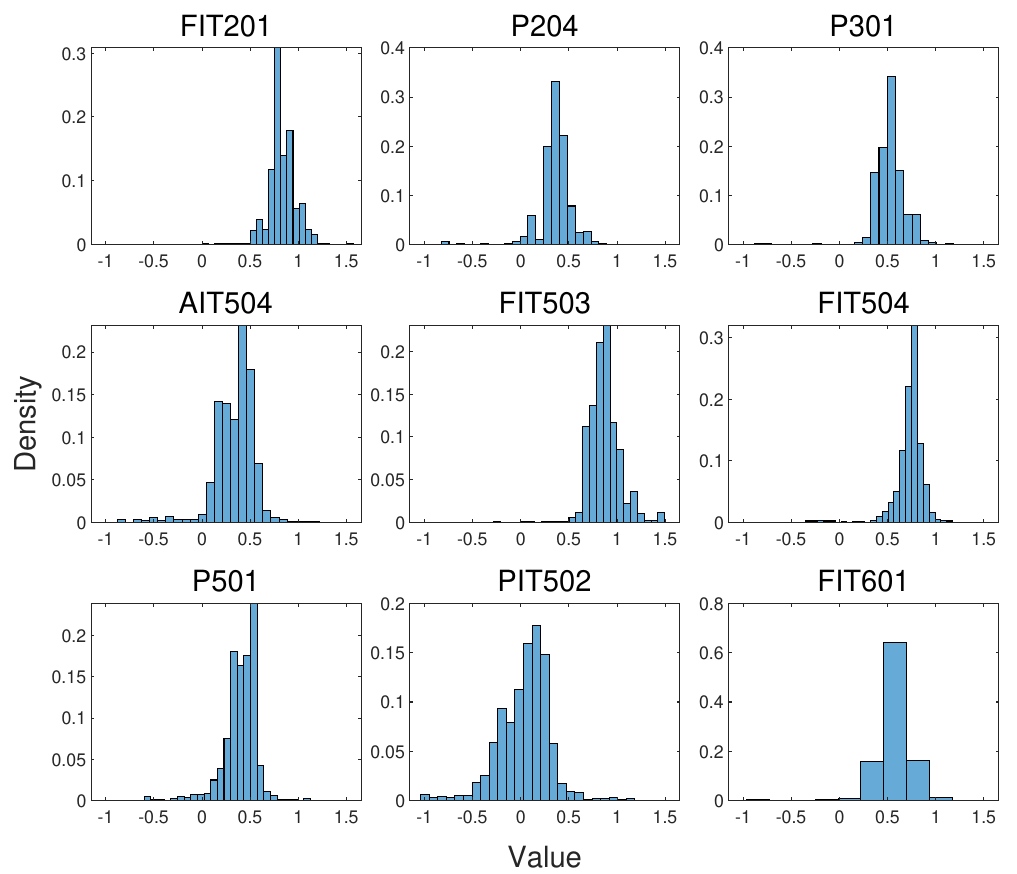}
    \caption{Transformed distributions of multiple entities.}
    \label{fig: Transformed distribution}
\end{figure}

\begin{figure*}
    \centering
    \includegraphics[width=1\textwidth]{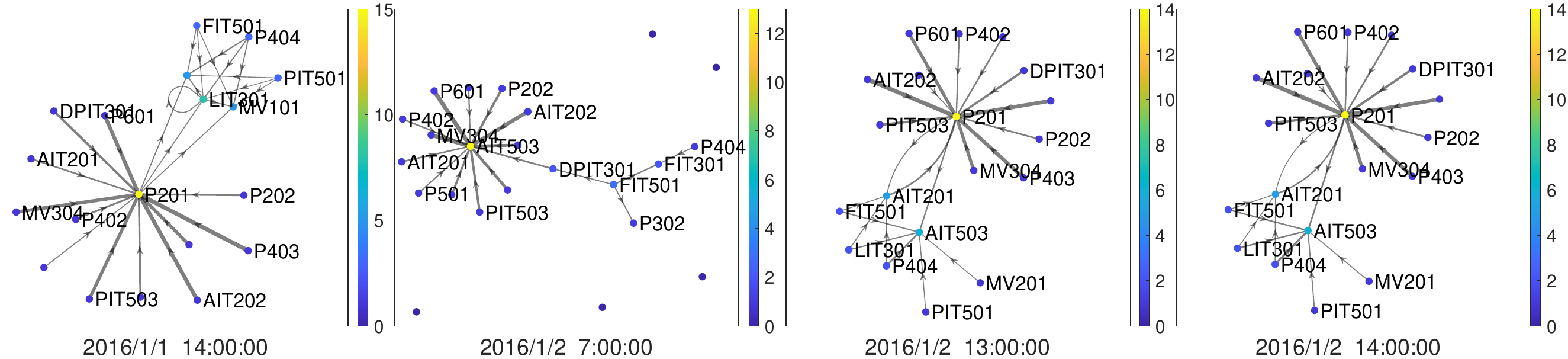}
    \caption{Dynamic graph structure in MTGFlow.}
    \label{fig: dynamic graph structure}
\end{figure*}

\subsubsection{Anomaly Ratio Analysis}
To further investigate the influence of anomaly contamination rates, we vary training splits to adjust anomalous contamination rates. For all the above-mentioned datasets, the training split increases from 60\% to 80\% with a 5\% stride. We present an average result over five runs in Fig.~\ref{fig: anomaly_ratio}. Taking SWaT as an example, the corresponding anomaly ratios of training datasets on SWaT are 17.7\%, 16.7\%, 15.7\%, 14.6\%, and 13.9\%, respectively. The detection performance of MTGFlow fluctuates in a small range, which shows MTGFlow is robust to the degree of anomaly contamination ratio. The same phenomenon can be seen in the rest datasets. Consequently, we can conclude that the anomaly detection performance of MTGFlow remains at a stable high level as the anomaly contamination ratio of the training dataset rises.

\subsection{Result Analysis}
\label{Result Analysis}
In order to further investigate the effectiveness of MTGFlow, we give a detailed analysis based on SWaT dataset.

\subsubsection{Entity-specific Density Estimation} 
We further explore whether the distributions of all entities are transformed into different target distributions to verify our entity-aware design. Since the window size is 60, the corresponding transformed distributions are also 60-dimensional distributions. Every single dimension of the multivariate Gaussian distribution is a Gaussian distribution. For better visualization, we present the 0-$th$ dimension of the transformed distributions across the whole test dataset in Fig.~\ref{fig: Transformed distribution}. Nine distributions of different entities are displayed. It can be seen that these distributions have been projected as unique distributions. Moreover, these distributions are successfully converted to preset Gaussian distributions with different mean vectors. The one-to-one mapping models entity-specific distributions and captures their respective sparse characteristics of anomalies, thus mitigating the above-mentioned
performance sacrifice.

\subsubsection{Dynamic Graph Structure}
Interdependencies among entities are not guaranteed to be immutable.
In fact, pairwise relations evolve with time. Benefiting from self-attention, MTGFlow can model this characteristic into a dynamic graph structure. We treat the attention matrix as the graph adjacent matrix. An empirical threshold of 0.15 is set for the adjacency matrix to show an intuitive learned graph structure in the test split. In Fig.~\ref{fig: dynamic graph structure}, the node size represents its node degrees, 
the arrow direction represents the learned directed dependence and the arrow width indicates the weight of the corresponding interdependencies.
The graph structure at 2016/1/1 14:00:00 is centered on the sensor $P201$, while the edges in the graph have completely changed and the center has shifted from $P201$ to $AIT503$ at 2016/1/2 7:00:00. This alteration of the graph structure may result from changes in working condition of the water treatment plant.
Besides, two similar graph structures can be found at 2016/1/2 13:00:00 and 2016/1/2 14:00:00. This suggests that the graph structure will be consistent if the interdependencies remain unchanged over a period of time, possibly due to repetitive work patterns of entities.
In addition, the main pairwise relations (thick arrow) at 2016/1/1 14:00:00 are similar to the ones at 2016/1/2 14:00:00, both centered on $P201$. It indicates that the interdependencies on multiple sensors are periodic. We also find mutual interdependencies from learned graph structures, such as the edges between $P201$ and $AIT201$ at 2016/1/2 13:00:00.

We summarize the findings: 

\begin{enumerate}{}{}
\item{Dynamic interdependencies among multiple entities.} 
\item{Consistent interdependencies among multiple entities.} 
\item{Periodic interdependencies among multiple entities.} 
\item{Mutual interdependencies among multiple entities.}
\end{enumerate}{}{}
 Therefore, it is necessary to use a dynamic graph to model such changeable interdependencies.

\begin{figure}[t]
  \centering
    \includegraphics[width=1\columnwidth]{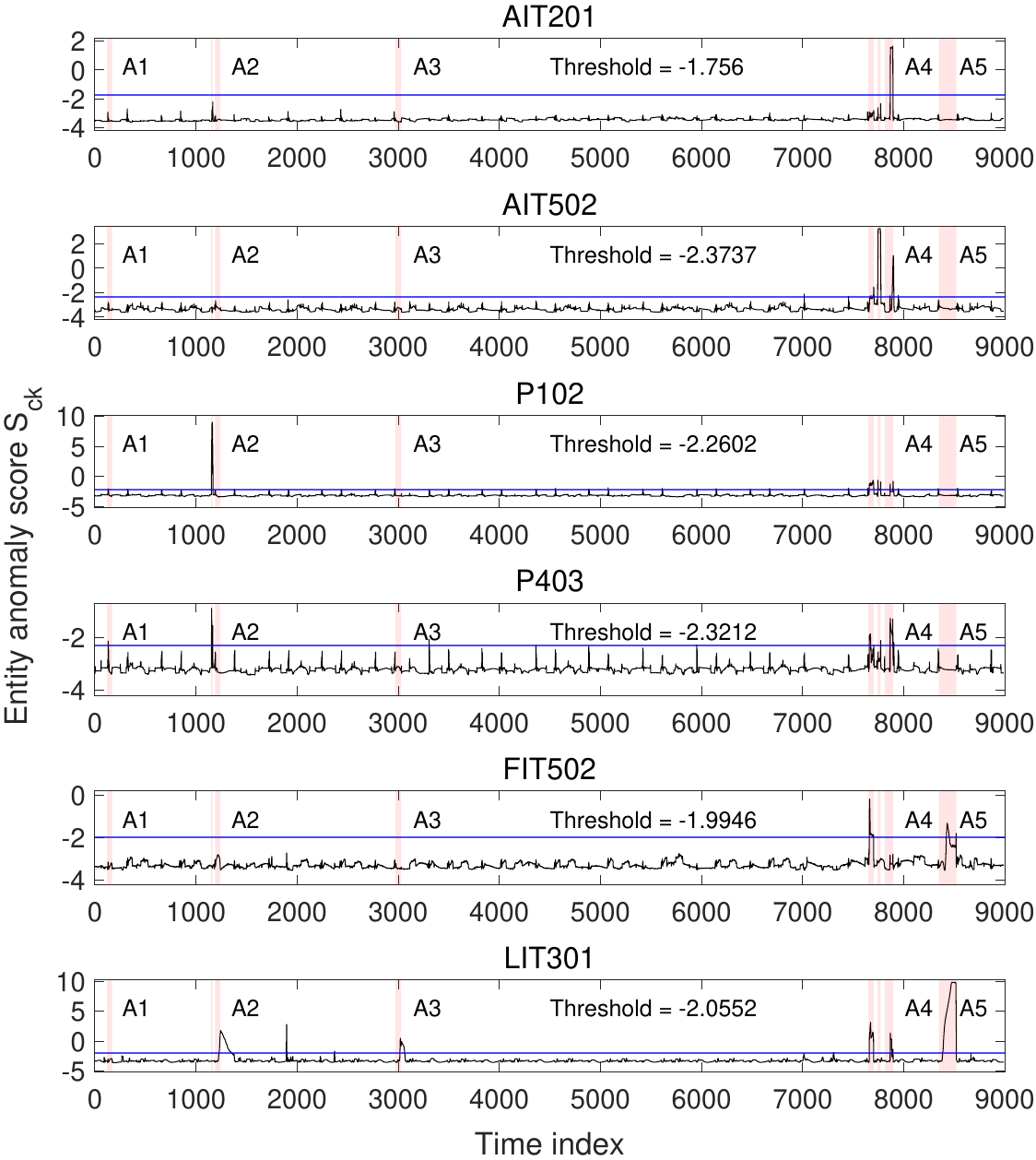}%

 \caption{Variation of log likelihoods for different entities on the whole testing dataset (Anomalies are highlighted in red, and the blue line is the threshold according to $S_{ck}$).}
  \label{fig:sparse_characteristics}
\end{figure}

\subsubsection{Distinct Sparse Characteristics}

To demonstrate that the sparse characteristics vary with different entities, we study changes of $S_{ck}$ along time on SWaT. 
As shown in Fig.~\ref{fig:sparse_characteristics}, $S_{ck}$ of $AIT201$, $P102$, $FIT502$, $AIT502$, $P403$, and $LIT301$ are presented. The highlighted regions denote marked anomalies. For a better illustration, we divide the anomalous regions as $A_1$, $A_2$, $A_3$, $A_4$, and $A_5$ along the timeline. We can observe that different entities react to different anomalies because of their different work mechanisms. Specifically, $AIT201$ and $AIT502$ has obvious fluctuations at $A_4$, while $P102$ reacts to $A_2$. In addition, $FIT502$ is sensitive to $A_4$ and $A_5$, yet $P403$ 
has responses in the face of $A_2$ and $A_4$. But $LIT301$ is sensitive to $A_2$, $A_3$, $A_4$, and $A_5$. Nevertheless, MTGFlow is still able to accurately distinguish and detect these anomalies.

\vspace{-2mm}
\subsection{Experimental summary}
Without the requirement for any labeled data, MTGFlow and MTGFlow\_cluster have demonstrated their capabilities to identify anomalies by leveraging their intrinsic low data density in Sec.~\ref{Experiment}. We attribute the superior performance of  MTGFlow and MTGFlow\_cluster to the following parts:

\begin{itemize}
\item Individual anomalous time series from different entities present diverse sparse characteristics. Instead of mapping all these individual time series to the same latent space, we introduce an entity-aware normalizing flow to model entity-specific density estimation. This allows each entity to be associated with a unique target distribution, which results in an improvement in performance by 4.1\%AUROC on SWaT and 1.6\%AUROC on WADI.
\item Since interdependencies among entities are not immutable, they actually evolve over time. Rather than modeling such interdependencies as static, we propose a dynamic graph structure to capture these changeable interrelationships. This innovation achieves the performance improvement of 2.4\% on SWaT and 0.6\%AUROC on WADI.
\item In reality, the characteristics of entities are strongly connected to their positions in common scenarios. In contrast to the approach of mapping all entities to unique distributions like MTGFlow, MTGFlow\_cluster further proposes the cluster-aware normalizing flow to obtain the commonality of these characteristics while maintaining inter-cluster uniqueness. This cluster strategy empowers MTGFlow\_cluster to outperform MTGFlow by 1.0\%AUROC on SWaT and 1.1\%AUROC on WADI.
\end{itemize}
\vspace{-2mm}
\section{Conclusion}
In this work, we proposed MTGFlow and MTGFlow\_cluster, unsupervised anomaly detection approaches for MTS based on the dataset with absolute zero known label. They first model the distribution of training data and then estimate the density of test samples, which are identified anomalies when they lie in low-density regions of this distribution. Extensive experiments on real-world datasets demonstrate their superiority, even if there exists high anomaly contamination. MTGFlow and MTGFlow\_{cluster} achieve SOTA anomaly detection methods for MTS and univariate series in both unsupervised and OCC settings. The superior anomaly detection performance of MTGFlow and MTGFlow\_{cluster} is attributed to dynamic graph structure learning and entity/cluster-aware density estimation. In addition, we explore various interdependencies that exist between individual entities from the learned dynamic graph structure. And a detected anomaly can be understood and localized via entity anomaly scores. In the future, we plan to apply our model to more flow models and further improve the practicality of our approach. 



%



\vspace{-3mm}
\ifCLASSOPTIONcompsoc
  \section*{Acknowledgments}
\else
  \section*{Acknowledgment}
\fi

This work was supported by
the National Natural Science Foundation Program of China under Grant
U1909207 and Grant U21B2029.

\ifCLASSOPTIONcaptionsoff
  \newpage
\fi



%


{\small
\bibliographystyle{IEEEtran}
\bibliography{tkde}
}

%

\begin{IEEEbiography}[{\includegraphics[width=1in,height=1.25in,clip,keepaspectratio]{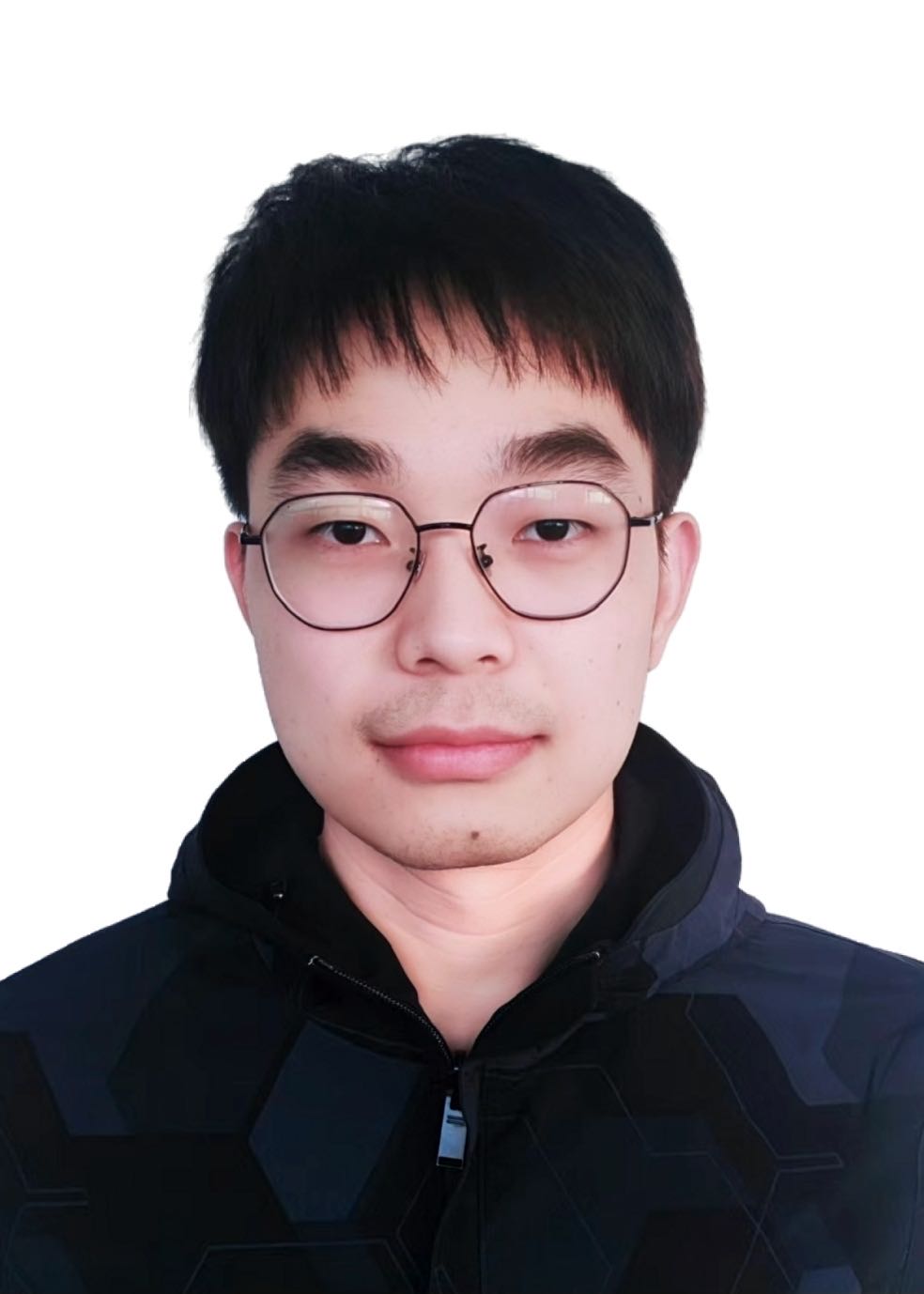}}]{Qihang Zhou}(Student Member, IEEE) received
the B.S. degree from China University of Geosciences, Wuhan, China, in 2020. He is currently pursuing the Ph.D. degree in control science
and engineering with the College of Control Science
and Engineering, Zhejiang University, Hangzhou,
China. His research interests include generative learning, data mining, and
anomaly detection.

\end{IEEEbiography}

\begin{IEEEbiography}[{\includegraphics[width=1in,height=1.25in,clip,keepaspectratio]{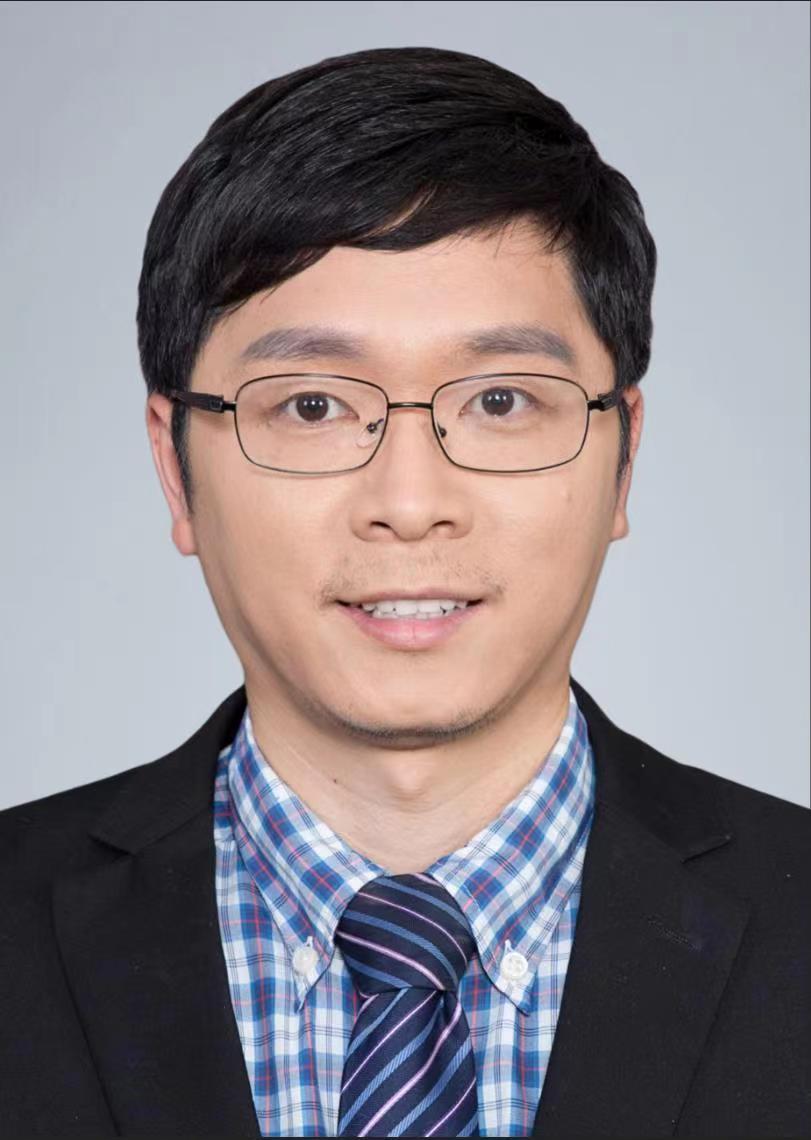}}]{Shibo He}
(Senior Member, IEEE) received the Ph.D. degree
in control science and engineering from Zhejiang
University, Hangzhou, China, in 2012. He is currently
a Professor with Zhejiang University. He
was an Associate Research Scientist from March
2014 to May 2014, and a Postdoctoral Scholar from
May 2012 to February 2014, with Arizona State
University, Tempe, AZ, USA. From November 2010
to November 2011, he was a Visiting Scholar with
the University of Waterloo, Waterloo, ON, Canada.
His research interests include Internet of Things,
crowdsensing, big data analysis, etc. Prof. He serves on the editorial board
for the IEEE TRANSACTIONS ON VEHICULAR TECHNOLOGY, Springer
Peer-to-Peer Networking and Application and KSII Transactions on Internet
and Information Systems, and is a Guest Editor for Elsevier Computer
Communications and Hindawi International Journal of Distributed Sensor
Networks. He was a General Co-chair for iSCI 2022, Symposium Co-Chair for the IEEE/CIC ICCC 2022, IEEE GlobeCom 2020
and the IEEE ICC 2017, TPC Co-Chair for i-Span 2018, a Finance and
Registration chair for ACM MobiHoc 2015, a TPC Co-Chair for the IEEE
ScalCom 2014, a TPC Vice Co-Chair for ANT 2013 and 2014, a Track Co-Chair
for the Pervasive Algorithms, Protocols, and Networks of EUSPN 2013, a
Web Co-Chair for the IEEE MASS 2013, and a Publicity Co-Chair of IEEE
WiSARN 2010, and FCN 2014.
\end{IEEEbiography}

\begin{IEEEbiography}[{\includegraphics[width=1in,height=1.25in,clip,keepaspectratio]{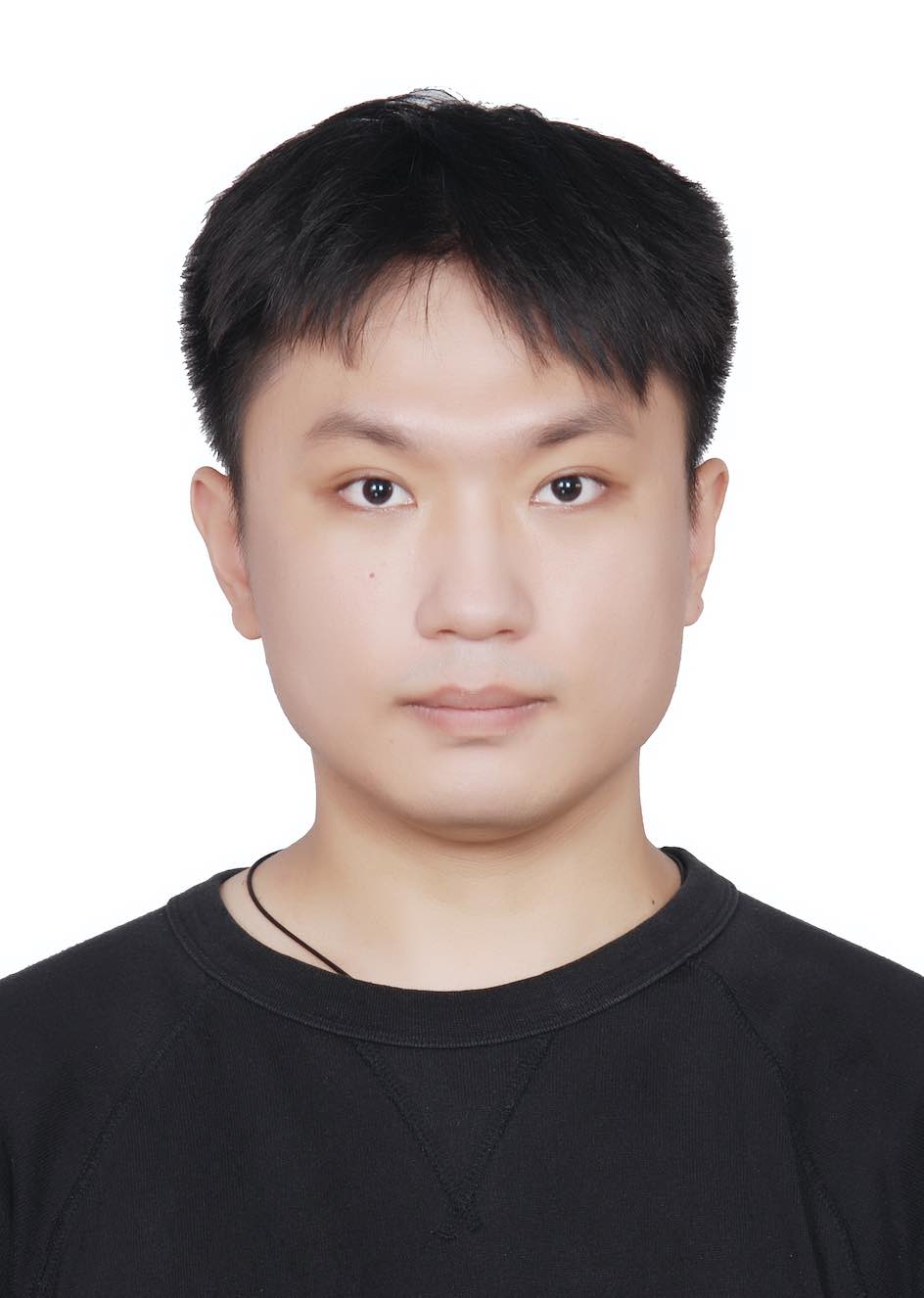}}]{Haoyu Liu}
received the Ph.D. degree from Zhejiang University, Hangzhou,
China, in 2021. He is currently working with
NetEase Fuxi AI Lab, Hangzhou. His research interests include data mining and machine learning, with particular interests in anomaly detection and crowdsourcing.
\end{IEEEbiography}

\begin{IEEEbiography}[{\includegraphics[width=1in,height=1.25in,clip,keepaspectratio]{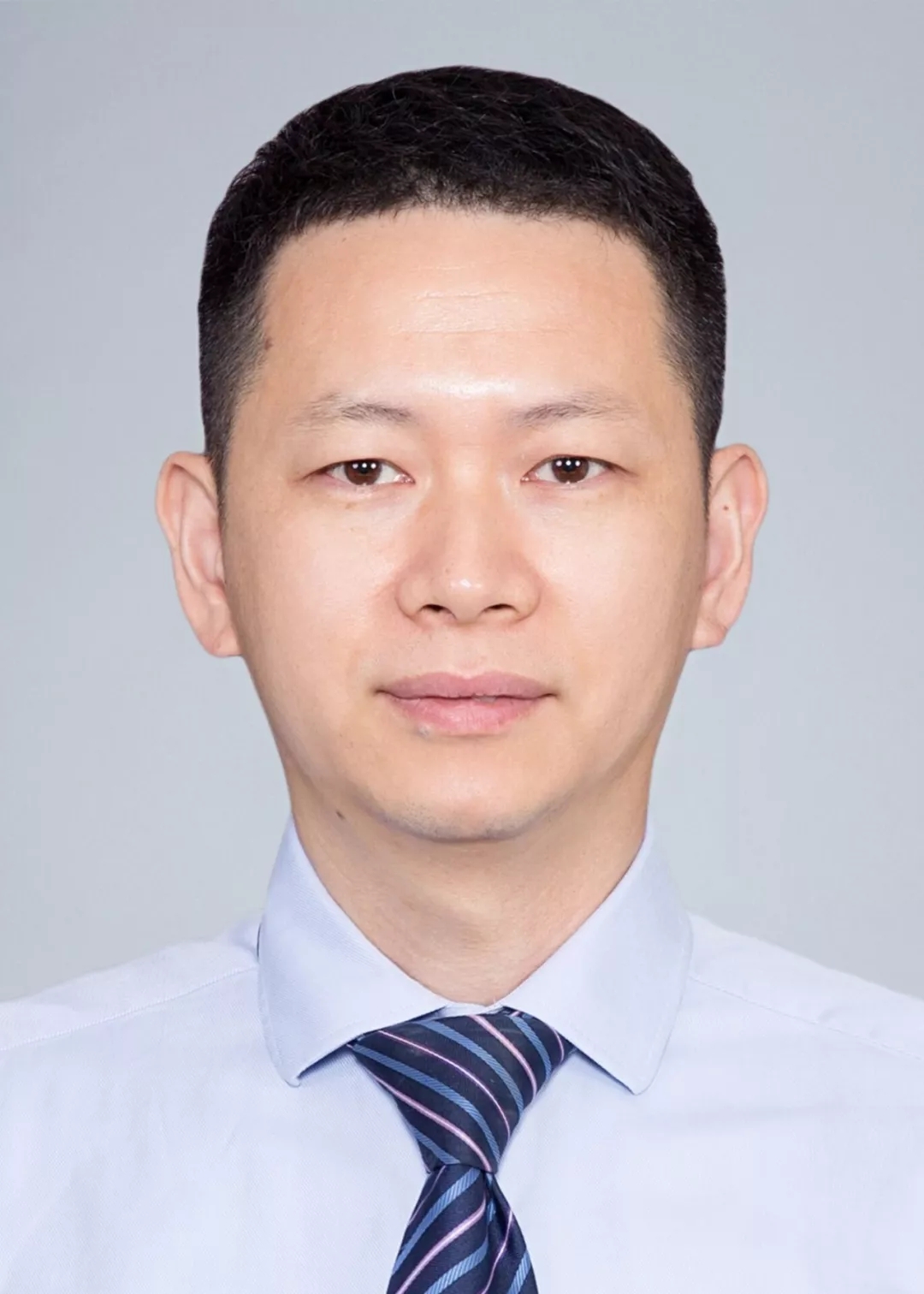}}]{Jiming Chen}
(Fellow, IEEE) received the PhD degree in control science and engineering from Zhejiang University, Hangzhou, China, in 2005. He is currently a professor with the Department of Control Science and Engineering, the vice dean of the Faculty of Information Technology, Zhejiang University. His research interests include IoT, networked control, wireless networks. He serves on the editorial boards of multiple IEEE Transactions, and the general co-chairs for IEEE RTCSA’19, IEEE Datacom’19 and IEEE PST’20. He was a recipient of the 7th IEEE ComSoc Asia/Pacific Outstanding Paper Award, the JSPS Invitation Fellowship, and the IEEE ComSoc AP Outstanding Young Researcher Award. He is an IEEE VTS distinguished lecturer. He is a fellow of the CAA.
\end{IEEEbiography}

\begin{IEEEbiography}[{\includegraphics[width=1in,height=1.25in,clip,keepaspectratio]{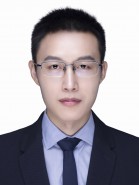}}]{Wenchao Meng}
(Senior Member, IEEE) received
the Ph.D. degree in control science and engineering from Zhejiang University, Hangzhou, China, in 2015, where he is currently with the College
of Control Science and Engineering. His current research interests include adaptive intelligent
control, cyber–physical systems, renewable energy
systems, and smart grids.
\end{IEEEbiography}








\end{document}